\documentclass[journal]{IEEEtran} 
\usepackage{times}

\usepackage[bookmarks=true]{hyperref}
\usepackage{booktabs} 
\usepackage[ruled,vlined,titlenotnumbered,linesnumbered]{algorithm2e} 
\usepackage{soul}
\SetAlFnt{\footnotesize}

\usepackage{graphicx,caption,subcaption}
\usepackage{amsmath,amssymb,stmaryrd,mathtools}
\usepackage[ruled,vlined,titlenotnumbered,linesnumbered]{algorithm2e} 
\usepackage[noend]{algpseudocode}
\usepackage{acronym}
\usepackage{comment}
\usepackage{color}
\usepackage{units}
\usepackage{color, colortbl}
\usepackage{adjustbox}
\usepackage{array,multirow}
\usepackage{xcolor}
\usepackage{soul}
\usepackage{wrapfig}

\definecolor{darkGray}{gray}{0.7}
\definecolor{lightGray}{gray}{0.9}

\usepackage[capitalise]{cleveref}

\DeclareCaptionLabelSeparator{periodspace}{.\quad}
\crefname{equation}{}{} 
\crefname{section}{Sec.}{Sec.}

\newcommand{\tvar}{t}
\newcommand{\thor}{0} 
\newcommand{\tinit}{0}

\newcommand{\ctrl}{u}
\newcommand{\dstb}{d}

\newcommand{\cset}{\mathcal{U}}

\newcommand{\dset}{\mathcal{D}}

\newcommand{\state}{\mathbf{x}}
\newcommand{\sset}{\mathcal{X}}
\newcommand{\traj}{\xi} 
\newcommand{\dyn}{f} 
\newcommand{\targetfunc}{l}
\newcommand{\targetset}{\mathcal{L}}

\newcommand{\costfunctionalBRT}{J}
\newcommand{\vfunc}{V}

\newcommand{\ham}{H}

\newcommand{\ctrlseq}{\mathbf{\ctrl}}
\newcommand{\dstbseq}{\mathbf{\dstb}}

\newcommand{\tdummy}{\tau}

\newcommand{\brt}{\mathcal{G}}
\newcommand{\brtlive}{\brt_{\text{live}}}
\newcommand{\brtunsafe}{\brt_{\text{unsafe}}}

\newcommand{\controller}{\pi}
\newcommand{\ctrlnom}{\pi_{\text{nom}}}

\newcommand{\horizon}{T}

\newcommand{\reals}{\mathbb{R}}

\newcommand{\livenessexample}{\textit{\textbf{Running Example (Liveness).~}}}
\newcommand{\safetyexample}{\textit{\textbf{Running Example (Safety).~}}}

\newtheorem{remark}{Remark}

\newtheorem{lemma}{Lemma}
\newtheorem{corollary}{Corollary}

\usepackage{xcolor,soul}
\soulregister\cite7
\soulregister\ref7
\soulregister\eqref7
\soulregister\aligned7
\colorlet{usercolorname}{red!00}
\sethlcolor{usercolorname}
\definecolor{softred}{rgb}{0.8, 0.2, 0.2}

\newif\ifshow 
\showfalse 

\ifshow
  \includecomment{wrap}
\else
  \excludecomment{wrap} 
\fi

\IEEEoverridecommandlockouts

\begin{document}
\title{\LARGE \bf On Safety and Liveness Filtering Using Hamilton-Jacobi Reachability Analysis}
\author{Javier Borquez, Kaustav Chakraborty, Hao Wang, Somil Bansal
\thanks{*Authors are with the ECE department at the University of Southern California. This research is supported in part by the DARPA ANSR program, the NSF CAREER program (award number 2240163), and Universidad de Santiago de Chile.
}
}
\maketitle

\begin{abstract}
Hamilton-Jacobi (HJ) reachability-based filtering provides a powerful framework to co-optimize performance and safety (or liveness) for autonomous systems.
Under this filtering scheme, a nominal controller is minimally modified to ensure system safety or liveness.
However, the resulting controllers can exhibit abrupt switching and bang-bang behavior, which is not suitable for applications of autonomous systems in the real world.
This work presents a novel, unifying framework to design safety and liveness filters through reachability analysis.
We explicitly characterize the maximal set of control inputs that ensures safety (or liveness) at a given state. 
Different safety filters can then be constructed using different subsets of this maximal set along with a projection operator to modify the nominal controller.
We use the proposed framework to design three safety filters, each balancing performance, computation time, and smoothness differently.
We highlight their relative strengths and limitations by applying these filters to autonomous navigation and rocket landing scenarios and on a physical robot testbed. 
We also discuss practical aspects associated with implementing these filters on real-world autonomous systems.
Our research advances the understanding and potential application of reachability-based controllers on real-world autonomous systems.

\end{abstract}


\section{Introduction} \label{sec:intro} 
As autonomous systems become integral to our daily lives, it is crucial to design controllers that are both safe and performant.
Typically, controller design is framed as an optimal control problem, balancing performance criteria (e.g., minimizing control energy), with safety constraints, such as obstacle avoidance. 
The resultant constrained optimal control problem is solved using a variety of methods ranging from Model Predictive Control (MPC) \cite{williams2016aggressive, darby2012mpc, borrelli2017predictive} and (approximate) dynamic programming \cite{bertsekas2012dynamic, bertsekas2019reinforcement} to data-driven, reinforcement learning approaches \cite{li2017deep, sutton2018reinforcement}.
Despite impressive performance of these methods, ensuring hard safety constraints for general, nonlinear systems remains a challenge, especially if the optimal control problem needs to be solved online. 

An alternative approach is \textit{safety filtering}, where a nominal control (often performance-oriented) is minimally adjusted (i.e., ``filtered'') to satisfy the safety constraint.
If the input is safe, it is used directly; otherwise, it is projected to a set of safe control inputs. 
This guarantees safety while optimizing performance whenever the system safety is not at risk.
Akin to safety filters, one can also design \textit{liveness filters}, which ensure that the system reaches a desired set of states (e.g., the landing pad for a rocket) within a set timeframe while optimizing the performance (e.g., minimizing the control energy).

A number of approaches have been proposed in literature to construct safety and liveness filters for dynamical systems, such as Control Barrier and Control Lyapunov functions, Hamilton-Jacobi reachability, and MPC.
We discuss some of the relevant approaches here and refer the interested readers to \cite{hsu2023safety} and \cite{data_driven_safety_filter} for a detailed survey on filtering methods.

Hamilton-Jacobi (HJ) reachability analysis \cite{Margellos11, mitchell2005time} is a popular mechanism to design these filters due to its ability to ensure safety and liveness for dynamical systems with general nonlinear dynamics, control bounds, and disturbances \cite{mitchell2004toolbox, bansal2017hamilton, deepreach}.
In reachability analysis, one is interested in computing the \textit{backward reachable tube (BRT)} of a dynamical system, i.e. the set of all initial states from which the system will eventually reach a target set, despite the worst case disturbance.
Thus, if the target set represents the set of desirable states, the BRT represents the set of states from which liveness can be guaranteed. 
Conversely, if the target set represents the set of undesirable states, the BRT contains states which are potentially unsafe for the system and should be avoided. 
Alongside the BRT, reachability analysis provides a safety controller (respectively liveness controller) that will provably steer the system away from the target set (respectively towards the target set).
A simple safety filter uses the BRT and safety controller: the system employs the nominal controller outside the BRT and switches to the reachability safety controller at the BRT boundary, ensuring constant safety. 
However, controllers from HJ reachability are often extremum seeking, resulting in a bang-bang behavior.
This behavior along with a sudden switching between the nominal and reachability controllers leads to jittery state and control trajectories, which is often undesirable for real-world autonomous systems \cite{data_driven_safety_filter}.

An alternative approach to filtering uses Control Barrier Functions (CBF) for safety and Control Lyapunov Functions (CLF) for liveness filtering.
Both utilize Lyapunov-like conditions to ensure the forward invariance of a set \cite{ames2017control}, which then aids in creating a quadratic program (QP) for a smooth blending of nominal and safety (or liveness) controllers.
However, despite recent progress \cite{cbvf, cbf_syn_svm, cbf_syn_expert_demo, cbf_syn_barrier_net}, constructing a valid Lyapunov or barrier function for general nonlinear systems with control bounds and disturbances remains a challenge.
Some recent studies have addressed this by framing CBF synthesis as an HJ reachability problem, capitalizing on the constructive attributes of reachability methods for obtaining a valid CBF \cite{cbvf}. 
The resultant CBF can be used for smooth, QP-based controller synthesis. 
However, a few questions remain: can we derive smooth control laws directly from reachability analysis without constructing a CBF first? 
How do these safety filters compare in performance and computation? 
And can we design such control laws for liveness filtering?


In this work, we answer the aforementioned questions in affirmative and introduce a general framework to design safety and liveness filters using reachability analysis.
Under the proposed framework, any safety and liveness filter can be thought of as a combination of two components: (a) a \textit{projection set} -- a set of safe or live control inputs that the nominal controller will be projected to; and (b) a \textit{projection operator} to align a nominal control input with this set.
Drawing on this insight, our key idea is to use HJ reachability analysis to explicitly characterize the set of \textit{all} possible control inputs at a state that guarantees system safety or liveness.
Various safety filters can then be 
obtained by using different subsets of this maximal safe control set as the projection set, along with a projection operator, to mathematically characterize the proposed framework we will use the notation presented in Table~\ref{tab:my_table}.
%
%
Building upon this approach, we propose three different safety and liveness filters based on reachability analysis: a least restrictive safety filter, a smooth least restrictive filter, and a smooth blending filter.
The proposed filters are versatile, applicable to both liveness and safety, accommodating control bounds, adversarial disturbances, model uncertainties, and time-specific liveness and safety properties (e.g., reach the target set within 5 seconds).
We further compare these filters on various metrics including performance, computation time, controller tuning, and control profile smoothness, as well as  a comparison to CBF methods for the safety filter.
Beyond these specific filters, our approach provides a unifying framework for existing reachability-based filters and designing new ones using diverse subsets of the maximal safe set and projection operators.
We showcase our approach in two different applications inspired by rocket landing within a given landing window and safe autonomous blimp navigation, as well as on a physical robot testbed involving a wheeled robot navigating through a cluttered environment.

\section{Problem Statement} \label{sec:formulation}
%
%
Consider an autonomous system with state $\state \in \sset \subseteq \mathbb{R}^n$ that evolves according to dynamics  
\begin{equation}\label{eq:sys_dyn}
    \dot{\state} = \dyn(\state, \ctrl, \dstb)
\end{equation}
where $\ctrl \in \cset$ and $\dstb \in \dset$ are the control and disturbance of the system, respectively. 
$\dstb$ can represent potential model uncertainties or an actual, adversarial exogenous input to the system.
We assume the dynamics are uniformly continuous in $u$ and $d$, bounded, and Lipschitz continuous in $\state$ for fixed $\ctrl$ and $\dstb$.
Finally, let $\traj_{\state,\tvar}^{\ctrlseq,\mathbf{\dstb}}(\tdummy)$ denote the system state at time $\tdummy$, starting from the state $\state$ at time $\tvar$ under control signal $\ctrlseq(\cdot)$ and disturbance signal $\dstbseq(\cdot)$ while following the dynamics \eqref{eq:sys_dyn}. A control signal $\ctrlseq(\cdot)$ is defined as a measurable function mapping from the time horizon to the set of admissible controls $\cset$, and a disturbance signal is similarly defined. 
\hl{We additionally assume that the control and disturbance signals $\ctrlseq(\cdot)$ and $\dstbseq(\cdot)$ are piecewise continuous in $t$. 
This assumption ensures that the system trajectory $\traj_{\state,\tvar}^{\ctrlseq,\mathbf{\dstb}}$
exists and is unique and continuous for all initial states \cite{coddington1955theory, callier2012linear}.} 

In this work, we are interested in
synthesizing potentially time-varying controllers $\controller: \left[\thor, \horizon\right] \times \sset \rightarrow \cset$ that steer the system to reach (liveness) or avoid (safety) a given target set $\targetset \subseteq \sset$, within the time horizon $\left[\thor, \horizon\right]$.
$\targetset$ can represent the goal region in the case of liveness or an unsafe region of the state space in the case of safety, e.g., obstacles for a navigation robot. 
Furthermore, we would like our controllers to consider other performance objectives while guaranteeing liveness/safety. 
We assume the performance objectives are encoded and optimized by an user-defined nominal controller $\ctrlnom$.
However, $\ctrlnom$ may not necessarily ensure liveness (or safety). 
\textit{Our goal is to design $\controller$ to follow $\ctrlnom$ to the extent possible, while guaranteeing liveness/safety requirement. }

One popular mechanism to achieve this goal is \textit{liveness or safety filtering}, in which $\ctrlnom$ is modified minimally to ensure system liveness or safety \cite{ames2019cbf, bansal2017hamilton}. 
In this paper, we will leverage reachability analysis to synthesize such filters.\\

\begin{table}[t!]
\centering
\begin{tabular}{|p{0.14\linewidth}  |p{0.75\linewidth}|} \hline 
\textbf{Symbol} & \textbf{Definition}   \\ \hline
$n\in\mathbb{Z}_+$ & Number of states \\  
$\sset\subseteq\reals^n$ & State space \\ 
$\state\in\sset$      & System state      \\ 
$\dot{\state}$ & Time derivative of state $\state$\\ 
$\cset$ & Control space \\  
$\ctrl\in\cset$       & Control input  \\  
$\dset$ & Disturbance space \\  
$\dstb\in\dset$       & Disturbance input \\ 
$f$ & System dynamics \\
$\ctrlseq(\cdot)$ & Control signal (over time)\\
$\dstbseq(\cdot)$ & Disturbance signal (over time) \\ 
$\traj_{\state,\tvar}^{\ctrlseq,\mathbf{\dstb}}(\tau)$ & State at time $\tau$ starting from state $\state$ at time $\tvar$ under control and disturbance signals $\ctrlseq(\cdot)$ and $\dstbseq(\cdot)$\\
$\targetset\subseteq\sset$ & Target set. Could represent goal or failure states.\\ 
$\targetfunc:\sset\!\rightarrow\!\reals$ & Target function. Signed distance to the target set. \\ 
$\horizon$ & Time horizon. \\
$\pi_{\text{nom}}$ & Nominal controller \\
$\vfunc(\state,\tvar)$ & Value function at state $\state$ and time $\tvar$ \\ 
$ D_{\tvar} \vfunc(\state, \tvar)$ & Time derivative of the value function \\
$\nabla \vfunc(\state, \tvar)$ & Spatial derivative of the value function\\
$\brtlive(\tvar)$ & BRT of the target set in the liveness problem. \\
$\controller_{\text{live}}^{*}(\state, \tvar)$ & Default liveness controller from the reachability analysis \\
$\cset_{\text {live}}(\state, \tvar)$ & Set of liveness preserving controls at state $\state$ and time $\tvar$. \\
$h$ & Projection operator for filtering \\
$\widetilde{\cset}$ & Projection set for filtering\\ 
$\gamma$ & Constant used in the smooth blending filter. \\
$\brtunsafe$ & BRT of the target set in the safety problem. \\
$\controller_{\text{safe}}^{*}(\state)$ & Default safety controller from the reachability analysis.\\ 
$\cset_{\text{safe}}(\state)$ & Set of safety preserving controls. \\ \hline
\end{tabular}
\caption{A summary of all the symbols used in the paper.}
\label{tab:my_table}
\end{table}

\noindent \livenessexample
We use a rocket landing system as a running example throughout this paper to demonstrate the liveness properties of our controllers. 
The rocket is modeled as a 6D system with dynamics,
\begin{equation}
\begin{aligned}
\label{eqn:dyn_liveness_running}
    \dyn_{\text{rocket}}(\state,\ctrl) = 
    \dfrac{d}{dt}\begin{bmatrix}
      y \\
      z  \\
      \theta   \\
     \dot y  \\ 
     \dot z  \\
     \dot \theta 
     \end{bmatrix} = 
     \begin{bmatrix}
     \dot y \\
     \dot z \\
     \dot \theta \\
     \cos(\theta)u_y - \sin(\theta)u_z  \\
     \sin(\theta)u_y + \cos(\theta)u_z - g \\
     \alpha u_1
     \end{bmatrix}
\end{aligned}
\end{equation}
Here $y$, $z$ denotes the horizontal and vertical positions of the center of mass of the rocket, $\theta$ denotes the rocket's heading, \hl{$u_y$ and $u_z$ denote the thrust in the $y$ and $z$ direction, respectively}, and $g$ is the acceleration due to the gravity of Earth. The task of the liveness controller is to generate thrust inputs $u = [|u_y| < 250, |u_z| < 250]$  such that the system can reach a landing pad (target set) within one second. \hl{The target set is defined as the rectangular area $\targetset = \{(y,z): |y|<20, 0<z<20\}$. The corresponding target function is given by $l(\state)=\max\{|y|-20, z-20, -z\}$, which approximates the signed distance to the target set}. 
For this example, we assume no external disturbances are acting on the system.

We first synthesize a nominal controller that attempts to take the system to $\targetset$.
The nominal controller is a sampling-based model predictive controller (MPC), whose objective is to minimize the distance between center of mass position and the boundary of $\targetset$, and the control energy over the entire trajectory.
Intuitively, the nominal controller can be considered as a performance control that tends to drive the system state to $\targetset$ while optimizing the amount of control effort.
\hl{The MPC cost to be minimized over the entire trajectory is given as,}
\begin{equation}
\begin{aligned}
\min_{u} \quad & \sum_{i=1}^{t}||\left[u_{y_{i}}, u_{z_{i}}\right]||_{2}+\sqrt{(|y_{i}|-20)^2+(|z_{i}|-20)^2}\\
\end{aligned}
\label{eqn:mpc_cost_liveness}
\end{equation}


\hl{Here, the first term penalizes the control cost while the second term penalizes the distance from the target set. $i$ is the discrete timestep and $||.||_2$ is the $L_2$ norm.}
The MPC controller is implemented in a receding horizon fashion with a horizon of 1000 steps. We use a first-order Euler discretization of the continuous dynamics in \eqref{eqn:dyn_liveness_running} with timestep, $\delta$ = 0.0005s. In our simulation studies, we will show that, in this case, such a nominal controller is insufficient to assure the liveness of our system.
Liveness might be achievable using MPC, but designing such a controller would require more care and time.
In this work, we will use a liveness filter to refine this nominal controller and produce a controller with the desired liveness properties. 

\vspace{0.3em}


\section{Background: Hamilton-Jacobi Reachability} \label{sec:reachability}
%
%
Hamilton-Jacobi (HJ) reachability analysis is a formal verification technique that characterizes the set of states from which the liveness (or safety) constraints can be satisfied with some control under the worst case disturbance.
In our work, we will leverage this technique to synthesize liveness (and safety) filters for the nominal controller.
In this section, we provide a brief overview of HJ reachability analysis. 

\subsection{Liveness and Safety Problems in HJ Reachability}
In HJ reachability analysis, the liveness and safety problems are posed as optimal control problems; informally, both problems intend to find a control signal $\ctrlseq(\cdot)$ that steers the system as ``deep'' into or as ``far away'' from the target set as possible.
To capture this semantic, the minimum distance between the system and the target set $\targetset$ over the time horizon is defined as the objective of the optimal control problems. Let $\targetfunc: \sset \rightarrow \mathbb{R}$ be some bounded and Lipschitz continuous function whose sub-zero level is given by the target set: $\targetset = \{\state: \targetfunc(\state) \leq 0 \}$.
Here, we present the reachability analysis for the liveness case and then comment on the safety case.
Given $\targetfunc$, the liveness problem is defined in \eqref{prob:liveness}:
%
\begin{equation} \label{prob:liveness}
\begin{split} 
    \max_{\dstbseq(\cdot)} \min_{\ctrlseq(\cdot)}& \min_{\tdummy \in [\tvar,\horizon]} \targetfunc(\traj_{\state,\tvar}^{\ctrlseq,\mathbf{\dstb}}(\tdummy)) \\
    s.t.     & \quad \dot{\state} = \dyn(\state, \ctrl, \dstb)
\end{split}
\end{equation}
%
%
\hl{where} the minimum cost over time for the trajectory is defined as the cost function $\costfunctionalBRT(\state, \tvar, \ctrlseq(\cdot),\dstbseq(\cdot)) = \min_{\tdummy \in [\tvar,\horizon]} \targetfunc(\traj_{\state,\tvar}^{\ctrlseq,\mathbf{\dstb}}(\tdummy))$.
The above optimization problem \eqref{prob:liveness} finds the minimum distance to the target set over the system trajectory, under the optimal control and the worst case disturbance. Thus, the system can reach the target set if and only if the minimum distance is less than 0, and this minimum distance is captured by the value function:
\begin{equation} \label{eq:vfunc_live}
    \vfunc(\state,\tvar) = \max_{\dstbseq(\cdot)} \min_{\ctrlseq(\cdot)}  \costfunctionalBRT(\state, \tvar, \ctrlseq(\cdot),\mathbf{\dstb}(\cdot))
\end{equation}
%
%

The value function \eqref{eq:vfunc_live} can be computed using dynamic programming, and it satisfies the following
final value \hl{Hamilton-Jacobi-Isaacs} Variational Inequality (HJI-VI) \cite{mitchell2005time, lygeros2004reachability}:
\begin{equation} \label{eq:HJIVI_BRS}
    \begin{aligned}
    \min \{D_{\tvar} \vfunc(\state, \tvar) &+ \ham(\state, \tvar, \nabla \vfunc(\state, \tvar)), \targetfunc(\state) - \vfunc(\state, \tvar) \} = 0 \\
    &\text{for} \ \tvar \in \left[\tinit, \horizon\right] \;
     \text{and} \;    \vfunc(\state, \horizon) = \targetfunc(\state) \quad 
    \end{aligned}
\end{equation}
Here, $D_{\tvar} \vfunc(\state, \tvar)$ and $\nabla \vfunc(\state, \tvar)$ denote the temporal derivative and the spatial gradients of the value function $\vfunc(\state, \tvar)$, respectively. 
The Hamiltonian encodes how the control and disturbance interact with the system dynamics and is given as:
%
\begin{equation}\label{eq:HJIVI_ham_live}
    \ham(\state, \tvar, \nabla \vfunc(\state, \tvar)) = \min_{\ctrl \in \cset} \max_{\dstb \in \dset} \nabla \vfunc(\state, \tvar) \cdot \dyn(\state, \ctrl, \dstb)
\end{equation}
%
%

Given the value function, one can also obtain the Backward Reachable Tube $\brtlive(\tvar)$ of the system. Since $\brtlive(\tvar)$ is defined to be the set of initial states from which the system can reach the target set within the time horizon $(T-t)$ despite worst-case disturbance, $\brtlive(\tvar)$ is the sub-zero level set of the value function:
\begin{equation}
    \brtlive(\tvar) \coloneqq \{\state: \vfunc(\state, \tvar) \leq 0 \}
    \label{eqn:brt_live_value}
\end{equation}
Consequently, liveness is guaranteed as long as the system state is inside $\brtlive(\tvar)$.
Conversely, if the system state is outside $\brtlive(\tvar)$, the liveness can't be ensured despite the best control effort.
Thus, any live controller must maintain the system state within $\brtlive(\tvar)$ at all time $t$.

The HJ reachability analysis for the safety problem is similar to that for the liveness problem, except that the control tries to avoid entering $\targetset$. Thus, the role of control and disturbance is switched in \eqref{prob:liveness} and \eqref{eq:HJIVI_ham_live}.
The sub-zero level of the value function gives us the BRT $\brtunsafe$, which represents the set of all states from which the system is guaranteed to enter the target set (the unsafe region in this case), despite the best control effort.
Conversely, if the system state is outside the $\brtunsafe$, there exists a controller that will keep the system safe, despite worst-case disturbance.

\subsection{Default Liveness (Safety) Controller from HJ Reachability}
Along with the value function and the BRTs, reachability analysis also provides a liveness/safety controller for the system (referred to as the \textit{default liveness/safety reachability controller} from here on). At state $\state$ and time $\tvar$, the default liveness reachability controller is given as:
\begin{equation}
\label{eq:default_liveness_ctrl}
\controller_{\text{live}}^{*}(\state, \tvar) =  \arg \min_{\ctrl \in \cset} \max_{\dstb \in \dset} \nabla \vfunc(\state, \tvar) \cdot \dyn(\state, \ctrl, \dstb)
\end{equation}
Similarly, the optimal disturbance is given as:
\begin{equation}
\label{eq:default_liveness_dstb}
\dstb^{*}(\state, \tvar) =  \min_{\ctrl \in \cset} \arg \max_{\dstb \in \dset} \nabla \vfunc(\state, \tvar) \cdot \dyn(\state, \ctrl, \dstb)
\end{equation}
Intuitively, at any time step, $\controller_{\text{live}}^*(\state, t)$ tries to \textit{maximally} steer the system state towards the target set \hl{(we omit the state and time dependencies of the optimal controller and disturbance onwards for compactness)}. 
To understand this, recall that the value function for the \hl{liveness} problem \eqref{prob:liveness}, $\vfunc(\state, \tvar)$ represents the minimum cost the system can achieve over the time horizon $[\tvar, \horizon]$, starting from state $\state$ and time $\tvar$ under worst case disturbance. 
In the liveness problem, we would like the system to reach a state of lower value.
$\controller_{\text{live}}^*$ seeks to accomplish this objective by finding a control which steers the system in the direction of greatest decent on the value function (i.e. minimizing the dot product between $\dot{\state}$ and the spatial gradient of the value function $\nabla \vfunc$).
Moreover, it can be shown that as long as the system starts inside $\brtlive(\tvar)$, it is guaranteed to eventually reach the target set under $\controller_{\text{live}}^{*}$ despite the worst-case disturbance \cite{bansal2017hamilton}. 
This is analogous to the safety case, with the difference that the controller would try to steer the system away from the target set.

Even though the default reachability controllers respect the liveness/safety constraints, they do not consider other objectives of interests, such as minimizing control energy or tracking some nominal controller. 
Thus, a natural solution is to blend the default reachability controller with a nominal controller that takes these performance criterion into consideration while respecting safety/liveness constraints, which is the focus of this paper.

\begin{figure} [t]
    \centering
    \includegraphics[width=\columnwidth]{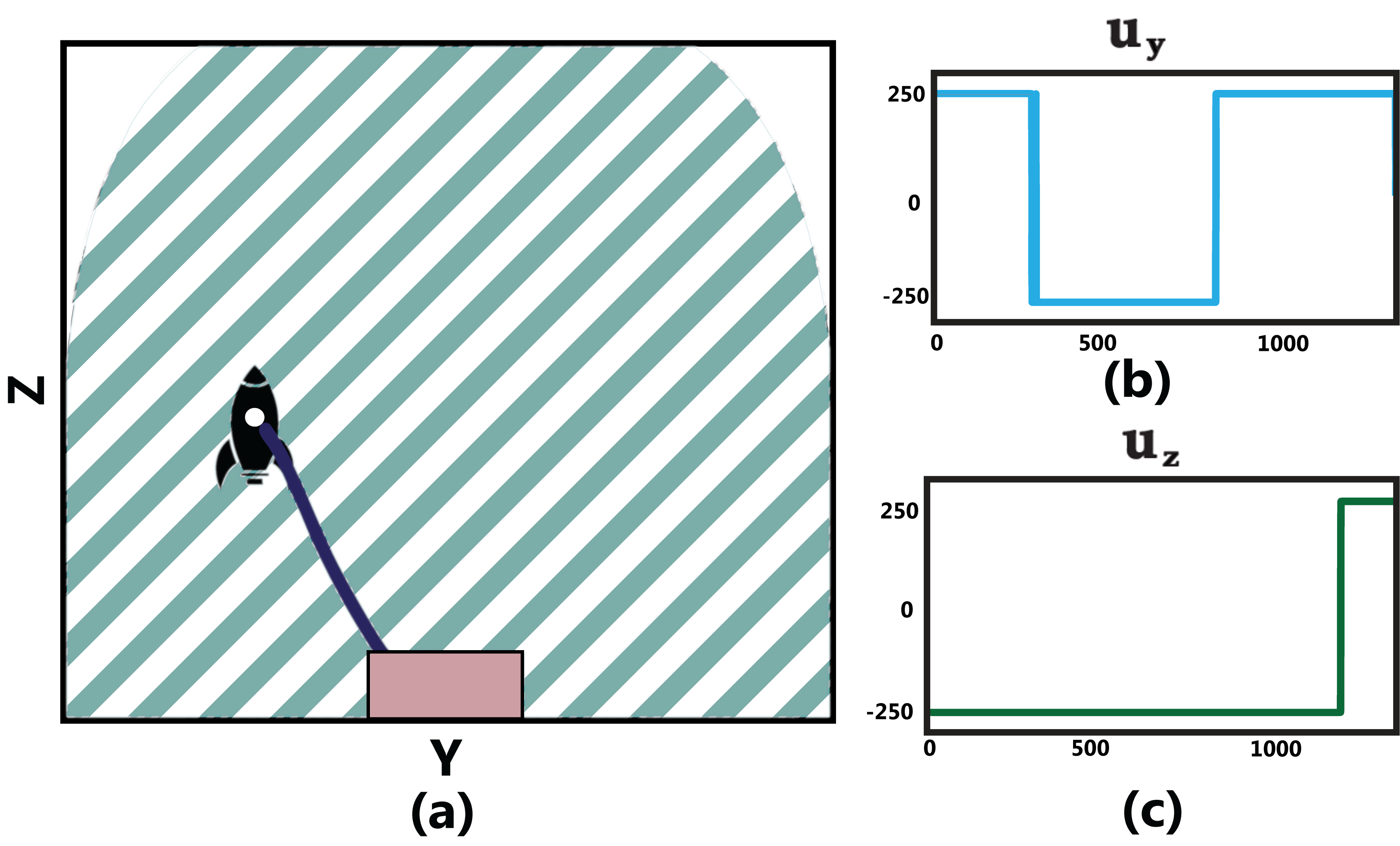}
    \caption{(a) BRT slice for a rocket landing problem in the $(y,z)$ states (all other states = 0) shown in teal. The target set $\targetset$ is shown in light pink rectangle. The trajectory followed by the rocket from its initial position (shown with the black rocket icon) under the default controller is shown in dark blue. (b) $u_y$ and (c) $u_z$
    controller profiles generated by the default reachability controller displaying bang-bang behavior.
    }\label{fig:brt_rocket_landing}
\end{figure}

\vspace{0.3em}
\noindent \livenessexample
As discussed earlier, in the liveness problem, we are interested in computing $\brtlive(\tvar)$, along with associated value function. 
%
This requires solving the HJI-VI \eqref{eq:HJIVI_BRS}.
Traditional methods compute a numerical solution of the HJI-VI over a state space grid \cite{mitchell2004toolbox}; however, these methods suffer from the curse of dimensionality and are not suitable for a 6D system.  
Instead, we use DeepReach \cite{deepreach} to compute the value function.
Rather than solving the HJI-VI over a grid, DeepReach represents the value function as a sinusoidal neural network and learn a parameterized approximation of the value function. 
Thus, memory and complexity requirements for training scale with the value function complexity
rather than the grid resolution. 
To train the neural network, DeepReach uses self-supervision on the HJI-VI itself. 
Ultimately, it takes as input a state $\state$ and time $\tvar$, and it outputs the value $\vfunc_{\beta}(\state, \tvar)$, where $\beta$ are the parameters of the NN.
We refer interested readers to \cite{deepreach} for further details.
For this system, we use a sinusoidal neural network with three hidden layers, and 512 neurons per layer. The training took approximately two hours for 20000 epochs on an NVIDIA RTX 4090 GPU. Using scenario optimization \cite{linScenario}, we verify the trained neural network by computing a high confidence error bound over the accuracy of the learned value function. 
A slice of the obtained BRT projected over the $YZ$-plane is shown in Fig. \ref{fig:brt_rocket_landing}.
As long as the system starts inside the teal region, it is guaranteed to reach the pink region (landing pad) under $\controller_{\text{live}}^{*}$.
The system trajectory under $\controller_{\text{live}}^{*}$ for one such starting state is shown in dark blue in Fig. \ref{fig:brt_rocket_landing}(a) and the corresponding control profiles are shown in Fig. \ref{fig:brt_rocket_landing}(b), (c).

As expected, the rocket eventually reaches the landing pad.
However, the default liveness controller only chooses the maximal control authority to steer the system (in this case, $|u_y| = |u_z| = 250$). 
This is a typical bang-bang nature of the reachability controllers for systems with control-affine dynamics and independently bounded control inputs, that ensures that the value function gradient is always pointing towards the direction of maximum decent.

However, as we will show in later sections, this approach results in jittery, high-energy control profiles that do not take into account any performance criteria, which results in a very high total cost for this control.

\section{Liveness Filtering Using HJ Reachability}
\label{sec:liveness}
The default controller obtained from the reachability analysis ensures liveness, but it does not consider other performance objectives.
In this section, we first use reachability analysis to characterize the set of \textit{all} liveness-ensuring control inputs at a particular state.
We then use this set to design different liveness filters.

\subsection{Characterizing the set of live controls}
At state $\state$ and time $\tvar$, the set of all liveness-ensuring control inputs (abbreviated as \emph{live controls} for the remainder of the paper) is given as:
\begin{align} \label{eqn:live_controller_set}
    \cset_{\text {live}}(\state, \tvar) &= \{u: \exists \epsilon > 0, \nonumber \\
    & V\left(\traj_{\state, \tvar}^{u, d^*}(t+\delta), t+\delta\right) \leq 0 ~~\forall \delta \in [0, \epsilon] \}
\end{align} 
Here, $V$ represents the value function obtained using the HJI-VI in \eqref{eq:HJIVI_BRS}.
$d^*$ is the optimal disturbance given by \eqref{eq:default_liveness_dstb}.
Intuitively, $\cset_{\text{live}}(\state, \tvar)$ \eqref{eqn:live_controller_set} is the set of all  controls that instantaneously keep the next state of the system inside (or at the boundary of) $\brtlive$, despite the worst case disturbance $d^*$.
This ensures that the system can still reach the target set from the next state, thereby ensuring liveness recursively.

We will next use \eqref{eqn:live_controller_set} to characterize $\cset_{\text {live}}(\state,\tvar)$. We assume that $\state \notin \targetset$; otherwise, the system state is already within the target set and no further control action is needed.

\begin{lemma}\label{lemma:u_live_set}
Assume that the value function $\vfunc(\state, \tvar)$ is differentiable for all $\state$ and $\tvar$.
Let $\state$ be any state such that $\state \notin \targetset$ and $\dstb^* \in \dset$ the optimal disturbance that maximally decreases liveness. Then the set of live controls at $\state$ is given as:
\begin{equation} \label{eqn:live_controls}
\cset_{\text {live}}(\state, t)=\left\{\begin{array}{l}
\cset \quad \text { if } \vfunc(\state, \tvar) < 0 \\
\\
\{u\in\cset:\\ D_{\tvar} \vfunc(\state, \tvar) + \nabla \vfunc(\state, \tvar) \cdot \dyn(\state, \ctrl, \dstb^*) = 0 \} \\
\quad \quad \quad \quad \quad \quad \text { if } \vfunc(\state, \tvar) = 0 \\
\\
\emptyset \quad \text { if } \vfunc(\state, \tvar) > 0 \end{array}\right.
\end{equation}
\end{lemma}
The proof of Lemma \ref{lemma:u_live_set} is in the Appendix.
Intuitively, \eqref{eqn:live_controls} states that $\cset_{\text{live}}(\state,\tvar)$ is the set of all admissible controls $\cset$ when the system state is in the interior of $\brtlive$.
This makes sense as the system can apply any control and instantaneously remain inside $\brtlive$ for a sufficiently small time step. 
On the other hand, when the system state is outside $\brtlive$, liveness can't be ensured despite the best control effort, resulting in an empty set. Finally, when the system state is at the boundary of $\brtlive$, $\cset_{\text{live}}(\state,\tvar)$ is given by the controls that ensure the \textit{total} derivative of the value function, $\frac{d}{dt}\vfunc(\state, \tvar)$, to be 0 (i.e., the control inputs that keep the system state instantaneously at the boundary of $\brtlive$). 

There are a few interesting observations to be made about the set of live controls in \eqref{eqn:live_controls}:
\begin{enumerate}
\item Lemma \ref{lemma:u_live_set} provides a time-dependent characterization of the live control inputs, allowing us to capture time-constrained liveness properties (e.g., reach the target set within $T$ seconds), as opposed to just asymptotic liveness. 
\item For control and disturbance-affine system dynamics, the condition of $\frac{d}{dt}\vfunc(\state, \tvar) = 0$ is linear in $u$. 
To see this, suppose the system dynamics are given by $f(\state,u,d) = f_1(\state)+f_2(\state)u +f_3(\state)d$. \hl{Its total derivative $\frac{d}{dt}\vfunc(\state, \tvar)$, under the optimal disturbance $\dstb^*$, is given by:}
\begin{align*}
& D_{\tvar} \vfunc(\state, \tvar) + \nabla \vfunc(\state, \tvar) \cdot \dyn(\state, \ctrl, \dstb^*) \\
&= D_{\tvar} \vfunc(\state, \tvar) + \nabla \vfunc(\state, \tvar) \cdot (f_1(\state)+f_2(\state)u +f_3(\state)d^*) \\
&= \alpha + \beta u
\end{align*}
We can observe that $\frac{d}{dt}\vfunc(\state, \tvar)$ is now a linear equation of $u$, where $\alpha = D_{\tvar} \vfunc(\state, \tvar) + \nabla \vfunc(\state, \tvar) \cdot (f_1(\state) + f_3(\state)d^*)$ and $\beta = \nabla \vfunc(\state, \tvar) \cdot f_2(\state)$. 
Thus, $\cset_{\text{live}}(\state,\tvar)$ is given by a hyperplane in $\cset$ (a point if the control space is one-dimensional), whenever the system state is at the boundary of $\brtlive(\tvar)$.
As we will see later, this will make the liveness filter design computationally efficient for control-affine systems.
\item Control inputs given by the default liveness controller \eqref{eq:default_liveness_ctrl} is always contained within $\cset_{\text{live}}(\state,\tvar)$.
The following corollary formalizes this result.
\begin{corollary}\label{cor:brt_live_u_live}
$\controller_{\text{live}}^{*}(\state,\tvar) \in \cset_{\text {live}}(\state, \tvar) \ \forall \state \in \brtlive$. 
\end{corollary}
Thus, the default controller can be used to steer the system to the target set, consistent with our expectation.
\end{enumerate}

%
The real utility of \eqref{eqn:live_controls} is that it can potentially provide more than one liveness-ensuring control inputs, whereas the default liveness controller can only provide one at a given state $\state$ and time $\tvar$.
We now formally establish that applying any control inputs from $\cset_{\text{live}}(\state,\tvar)$ is sufficient to ensure that the system will eventually reach the target set within the established time horizon $[t,T]$.
\begin{lemma}\label{lemma:u_live_guarantee}
\hl{Suppose the system starts inside $\brtlive$ at time $\tvar$. If the system applies control $u(\tdummy) \in \cset_{\text {live}}(\state(\tdummy), \tdummy) \ \forall \tdummy \in [\tvar, T]$, then $\exists s \in [\tvar, T]$ such that $\state(s) \in \targetset$.}
\end{lemma}

Lemma \ref{lemma:u_live_guarantee} effectively allows us to use any subset of $\cset_{\text{live}}(\state,\tvar)$ for the projection of the nominal controller, while still maintaining liveness. 
This suggests the following general structure for liveness filters:
\vspace{0.5em}

\noindent \textbf{General Liveness Filter}. Given a nominal controller, $\ctrlnom$, a general liveness filter can be formulated as:
\begin{equation} \label{eqn:general_liveness_filter}
    \begin{split} 
    \controller(\state, \tvar) = \arg\min_{\ctrl} & \quad h(\ctrl, \ctrlnom(\state, \tvar)) \\
    s.t. ~& \ctrl \in \widetilde{\cset}(\state, t), \text{~with~} \widetilde{\cset}(\state, t) \subseteq \cset_{\text{live}}(\state,\tvar)
\end{split}
\end{equation}
In \eqref{eqn:general_liveness_filter}, we refer to $h$ as the \textit{projection operator} and $\widetilde{\cset}$ as the \textit{projection set}.
Since $\widetilde{\cset} \subseteq \cset_{\text{live}}$, the above filter makes sure that $\controller(\state, \tvar) \in \cset_{\text{live}}(\state,\tvar)$, thereby ensuring system liveness at all times (by Lemma \ref{lemma:u_live_guarantee}).
Nevertheless, different choices of $h$ and \hl{$\widetilde{\cset}$} will lead to different trade-offs between performance and liveness.

To make sure that the optimization problem \eqref{eqn:general_liveness_filter} can be solved in a computationally efficient manner, it is often desirable to use a projection operator and set that are convex in $u$ for a given $\state$ and $t$.
A particularly popular choice in the literature for $h$ is the $l_2$ distance from the nominal controller, i.e., $h := ||\ctrl - \ctrlnom(\state, \tvar)||_2^2$.
For the remainder of this section, we use this $l_2$ projection operator and choose three different $\widetilde{\cset}$ that result in particularly interesting liveness filters.
We will conclude the section by comparing the resultant filters.

\subsection{Least Restrictive Filter}
To blend performance with liveness, the reachability analysis is typically applied in a least-restrictive fashion \cite{bansal2017hamilton}:
%
\begin{equation}\label{lst_restrict_liveness_ctrl}
\controller(\state, \tvar) = \begin{cases}
  \ctrlnom(\state, \tvar)&\quad \text { if } \vfunc(\state, \tvar)< 0 \\
   \controller_{\text{live}}^*(\state, \tvar) & \quad \text { if } \vfunc(\state) = 0
\end{cases}
\end{equation}
The \textit{least restrictive (LR) filter} follows the nominal controller when the system state $\state$ is in the interior of $\brtlive(\tvar)$ (i.e. $\vfunc(\state, \tvar) < 0$), and it takes corrective actions given by $\controller^*_{\text{live}}(\state,\tvar)$ when the system is on the boundary of or at the risk of exiting $\brtlive(\tvar)$. 
This ensures that the system never exits $\brtlive(\tvar)$.
The controller in \eqref{lst_restrict_liveness_ctrl} is least restrictive in the sense that it follows the given nominal controller that can optimize criteria besides liveness, and only interferes when the system is at the risk of breaching liveness. 
Consequently, the system can optimize other objectives while progressing towards the goal, but such freedom is not afforded by $\controller_{\text{live}}^*(\state, \tvar)$ as its sole objective is to ensure system's liveness. 

The LR filter \eqref{lst_restrict_liveness_ctrl} can be obtained using our general framework \eqref{eqn:general_liveness_filter} by using the following $\widetilde{\cset}$:
\begin{equation}
\widetilde{\cset}(\state, \tvar) = \begin{cases}
  \cset&\quad \text { if } \vfunc(\state, \tvar)< 0 \\
   \{\controller_{\text{live}}^*(\state, \tvar)\} & \quad \text { if } \vfunc(\state, \tvar) = 0
\end{cases}
\end{equation}
Thus, $\widetilde{\cset}$ is the set of all permissible controls whenever the system state is inside $\brtlive(\tvar)$ and is given by a singleton when it is at the BRT boundary.
It is easy to verify that $\widetilde{\cset}(\state, \tvar) \subset \cset_{\text{live}}(\state,\tvar)$ for all $\state$ and $t$. 
Thus, the system will always remain live under the LR filter.

\begin{figure}[b]
    \centering
\includegraphics[width=0.7\columnwidth]{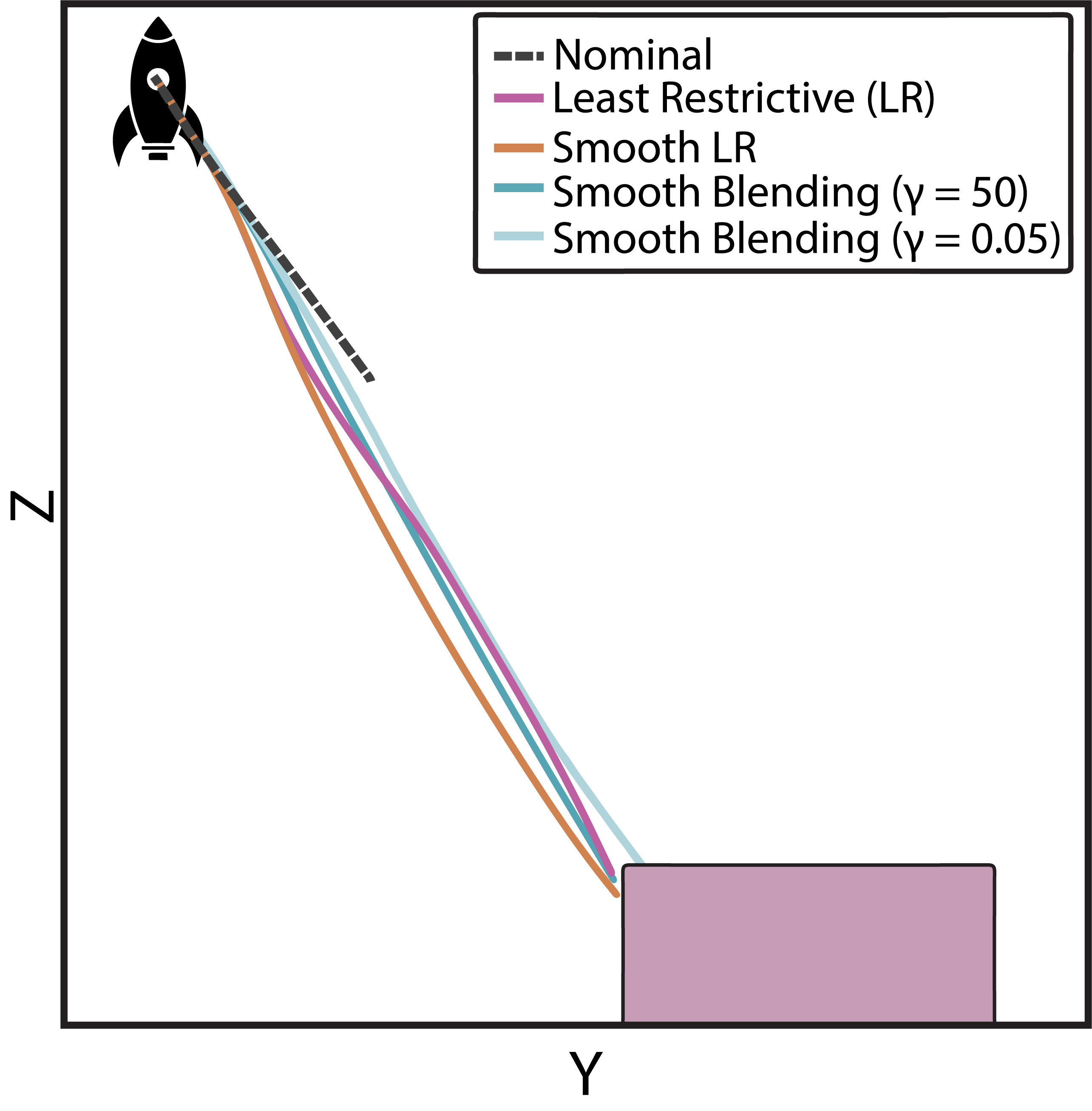}
    \caption{System trajectories under the nominal and filtered liveness controllers for the rocket landing system. \hl{Pink region represents the target set. 
    }}
    \label{fig:live_ctrl_traj}
    \vspace{0em}
\end{figure}
%

\begin{figure}[t]
    \centering
\includegraphics[width=1\columnwidth]{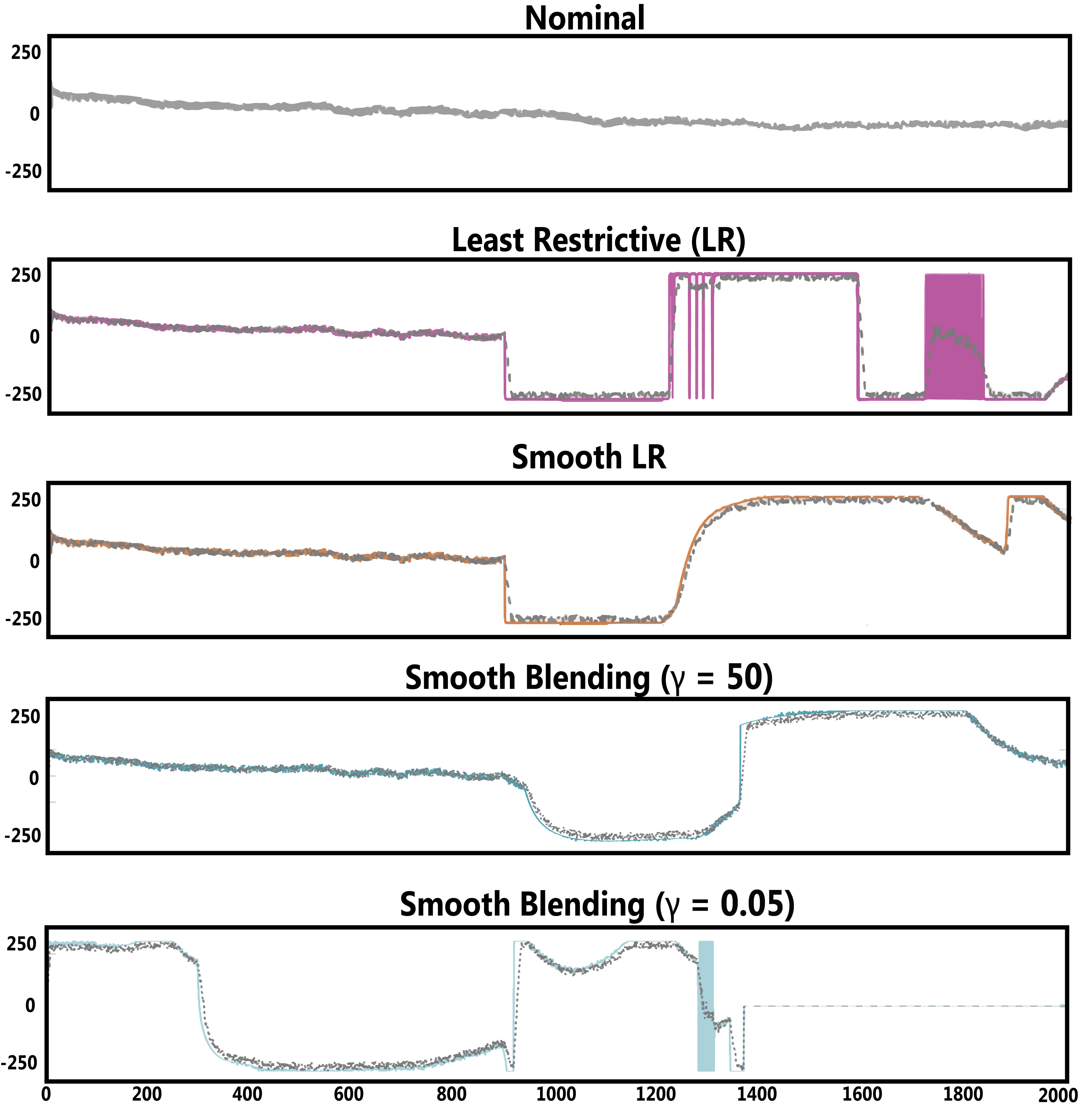}
    \caption{Control profiles for $u_1$ for different liveness filters. The nominal controller is shown in dashed gray.
    }
    \label{fig:liveness_ctrl_profiles_1}
    \vspace{0em}
\end{figure}

\vspace{0.3em}
\noindent \livenessexample
We now demonstrate the LR filter \eqref{lst_restrict_liveness_ctrl} on the running example.
The corresponding system trajectory and the control profile (for $u_1$) are shown in purple in Fig. \ref{fig:live_ctrl_traj} and \hl{Fig. \ref{fig:liveness_ctrl_profiles_1}}, respectively.
As expected, the LR filter chooses between $\controller^*_{\text{live}}$ and $\pi_{\text{nom}}$ to steer the system towards the target set.
However, as discussed earlier, using $\controller^*_{\text{live}}$ results in bang-bang behavior (as evident from the spikes in the control profile for the LR filter in Fig. \ref{fig:liveness_ctrl_profiles_1}) and high control energy. 
In the next subsection, we propose a smoother version of the LR filter to overcome this challenge.

\begin{remark}
Upon careful observation, one might note that the LR filter does not take the system fully inside the target set within the time horizon $T$ (the purple trajectory is slightly outside the target set in Fig. \ref{fig:live_ctrl_traj}).
This discrepancy between theory and practice is due to the effect of using a discrete-time simulation of a continuous-time system, which sometimes causes a slight delay in the switching between the nominal and the default controller. 
We will discuss this aspect more as well as a few potential solutions in Sec. \ref{sec:practical}.
\end{remark}
%

\subsection{Smooth Least Restrictive Filter}
While the LR filter \eqref{lst_restrict_liveness_ctrl} accounts for performance and liveness, it still switches to a high energy, bang-bang policy  $\controller^*_{\text{live}}$ at the boundary of $\brtlive$. 
To overcome this challenge, we propose an alternative liveness filter that uses the full control authority available to the system as the projection set, rather than just $\{\controller^*_{\text{live}}\}$:
%
\begin{equation}\label{opt:smooth_lst_res_ctrl_opt}
 \begin{split} 
    \controller(\state, \tvar) = \arg\min_{\ctrl \in \cset} & \quad ||\ctrl - \ctrlnom(\state, \tvar)||_2^2\\
     s.t.     & \quad \ctrl \in \cset_{\text{live}}(\state, \tvar)
 \end{split}
 \end{equation}
i.e., $\widetilde{\cset}(\state, \tvar) = \cset_{\text{live}}(\state,\tvar)$ for all $(\state, \tvar)$. Intuitively, the above optimization problem computes a control input at each time instant that is ``closest'' to the nominal controller, yet is within the set of live controls.
Furthermore, since $\cset_{live} = \cset$ whenever $\vfunc(\state, \tvar) < 0$, we can simplify the filter in \eqref{opt:smooth_lst_res_ctrl_opt} as:
\begin{equation}\label{ctrl:smooth_lst_res_ctrl}
\controller(\state, \tvar) = \begin{cases}
      \ctrlnom(\state, \tvar)& \vfunc(\state, \tvar)< 0 \\
       \controller^{+}(\state, \tvar) & \vfunc(\state, \tvar) = 0
    \end{cases}
\end{equation}
where $\controller^{+}(\state, \tvar)$ is obtained by solving the following optimization problem: 
\begin{equation}\label{opt:smooth_lst_res_ctrl_opt_simplified}
\begin{split} 
    \min_{\ctrl \in \cset} & \quad ||\ctrl - \ctrlnom(\state, \tvar)||_2^2 \\
    s.t.     & \quad D_{\tvar} \vfunc(\state, \tvar) + \nabla \vfunc(\state, \tvar) \cdot \dyn(\state, \ctrl, \dstb^*) = 0
\end{split}
\end{equation}
where $\dstb^*$ is the optimal disturbance given by \eqref{eq:default_liveness_dstb}.
We refer to the controller in \eqref{ctrl:smooth_lst_res_ctrl} as the \textit{smooth least restrictive (smooth LR) filter}.
Similar to the LR filter \eqref{lst_restrict_liveness_ctrl}, the smooth LR filter \eqref{ctrl:smooth_lst_res_ctrl} follows $\controller_{\text{nom}}$ when the system state is in the interior of $\brtlive(\tvar)$. 
However, when the system state is at the boundary of $\brtlive(\tvar)$, rather than using $\controller^*_{\text{live}}(\state,\tvar)$, it chooses a live control that is \textit{closest} to the nominal control by solving a Quadratic Program (QP), thus allowing for a smoother control profile.
\begin{remark}
The objective of the optimization problem \eqref{opt:smooth_lst_res_ctrl_opt_simplified} need not to be quadratic. 
As long as the objective is convex in $\ctrl$, the optimization problem in \eqref{opt:smooth_lst_res_ctrl_opt_simplified} remains convex in $\ctrl$ for control-affine systems and can be solved efficiently online. 
\end{remark}

\vspace{0.3em}
\noindent \livenessexample
The trajectory and control profile corresponding to the smooth LR filter are shown in orange in Fig. \ref{fig:live_ctrl_traj} and \ref{fig:liveness_ctrl_profiles_1}.
We observe that the smooth LR filter effectively ``smoothes'' out the bang-bang behaviors that the LR filter displayed; control profile now follows a gradual curve dictated by the QP formulation. This results in lower control energy and less jittery behavior of the resulting system when compared to the original LR controller.

\subsection{Smooth Blending of Performance and Liveness}
The LR and smooth LR filters employ live controls as a last-minute resort when the system is at the risk of breaching liveness. Despite its simplicity, this sudden and inconsistent switching to a liveness controller has a few drawbacks: (a) switching at the BRT boundary can still result in a jittery control profile if the nominal control is far from the set of live controls; (b) a last-minute switching can lead to liveness violation upon the slightest delay in switching.
To remedy these drawbacks, we take inspiration from the CBF literature. 
Specifically, we introduce a ``CBF-like'' constraint to define the projection set that encourages the filtered control to be ``more'' liveness-ensuring as the system approaches the boundary of $\brtlive(\tvar)$: 
%
\begin{equation} \label{ctrl:smooth_blend_liveness}
\begin{split} 
\widetilde{\cset}(\state, t) = & \{u \in \cset:\\
& D_{t} \vfunc(\state, t) + \nabla \vfunc(\state, t) \cdot \dyn(\state, \ctrl, \dstb^*) \leq -\gamma V(\state, t)\}
\end{split}
\end{equation}
\hl{where $\gamma$ is a tunable parameter that determines how quickly a control is permitted to drive the systems towards the boundary of $\brtlive(\tvar)$.} The left hand side of the constraint in \eqref{ctrl:smooth_blend_liveness} is the total derivative of $\vfunc(\state,\tvar)$ evaluated at state $\state$ and time $\tvar$.
Thus, intuitively, the above constraint limits the rate at which the value function can increase, i.e., how quickly is the system allowed to approach the boundary of $\brtlive(\tvar)$, at any given state $\state$. When the system is on the boundary of $\brtlive(\tvar)$, we have $-\gamma\vfunc(\state,\tvar)=0$, and the controller must output a control that would drive the system to a state that has same value. 
Specifically comparing with the LR and smooth LR filters in this section, the projection set in \eqref{ctrl:smooth_blend_liveness} blends $\pi_{\text{nom}}$ with $\controller^*_{\text{live}}$ in such a way that the system can take gradually more stringent corrective actions as the system becomes more at risk of breaching the liveness, resulting in less jerky controls.

The behavior of the control constraint in \eqref{ctrl:smooth_blend_liveness} is quite similar to that of the CBF constraint in CBF-QP \cite{ames2019cbf}, as $\gamma$ determines how quickly the value function can change. However, it can be difficult to find a valid CBF function in the presence of control bounds and disturbances, requiring online tuning of $\gamma$ to make sure that the constraint remains feasible at all times, which can be quite challenging in practice.
On the other hand, the reachability-based controller in \eqref{ctrl:smooth_blend_liveness} is always \textit{live} and \textit{feasible}, as we prove next, bypassing the need of tuning $\gamma$ to ensure liveness.
\begin{lemma} \label{lemma:feasibility_smooth_blending}
Let $\state$ be any state such that $V(\state, t) \leq 0$ and $\state \not\in \targetset$.
Define $\widetilde{\cset}(\state, t)=\{u \in \cset: D_{t} \vfunc(\state, t) + \nabla \vfunc(\state, t) \cdot \dyn(\state, \ctrl, \dstb^*) \leq -\gamma V(\state, t)\}$.
Then (a) $\widetilde{\cset}(\state, t)$ is non-empty, and (b) $\widetilde{\cset}(\state, t) \subseteq \cset_{\text{live}}(\state, t)$ for all $\gamma \geq 0$.
\end{lemma}
Intuitively, Lemma \ref{lemma:feasibility_smooth_blending} states that (a) using the projection set in \eqref{ctrl:smooth_blend_liveness} always results in a feasible filtering problem, and (b) liveness is always guaranteed under the resultant control input, regardless of the value of $\gamma$.
An alternative interpretation of Lemma \ref{lemma:feasibility_smooth_blending} is that, as long as the value function $V(\state, t)$ is differentiable, it  acts as a robust CBF for any value of $\gamma$ and can be used to synthesize CBF-like controllers.
    Even though a number of works have synthesized CBFs using the reachability analysis (e.g., \cite{cbvf}), these methods need to explicitly select a value of $\gamma$ during the value function computation, making it challenging to tune $\gamma$ online.
    In contrast, Lemma \ref{lemma:feasibility_smooth_blending} indicates that $\gamma$ can in fact be chosen online, independent of the value function computation. 
    
Nevertheless, the value of $\gamma$ still affects the blending of $\pi_{\text{nom}}$ and $\controller^*_{\text{live}}$.
Specifically, as $\gamma \rightarrow \infty$, the constraint in \eqref{ctrl:smooth_blend_liveness} no longer limits the rate of change in the value function and the blending controller starts behaving like the smooth LR filter \eqref{ctrl:smooth_lst_res_ctrl}. 
Thus, the system prioritizes the performance until the liveness is at risk. 
On the other hand, as $\gamma \rightarrow 0$, the constraint in \eqref{ctrl:smooth_blend_liveness} becomes very stringent and does not allow value function to increase.
Consequently, the blending controller starts behaving like the default liveness controller in \eqref{eq:default_liveness_ctrl}, prioritizing the system liveness over performance.
Thus, by choosing $\gamma$ in between the two extremes, we are effectively blending these two controllers, creating different trade-offs between liveness and performance.
%

\vspace{0.3em}
\noindent \livenessexample
We show the smooth blending filter for different values of $\gamma$. 
The trajectories are shown in different shades of cyan in Fig. \ref{fig:live_ctrl_traj}. 
In the first case with $\gamma = 0.05$, the system's behavior is very similar to the default reachability controller, where the system is very cautious and avoid moving towards the BRT boundary. 
However, from the corresponding control profile in Fig. \ref{fig:liveness_ctrl_profiles_1}, we notice that the controller follows a slightly smoother version of the bang-bang behavior that we saw in Fig. \ref{fig:brt_rocket_landing} and the control applied is not always the maximal control, especially towards the later times.
On the other hand, using $\gamma = 50$, the system goes very close to the boundary of the BRT, behaving similarly to the smooth LR filter, as expected.

\begin{figure} [t]
    \centering
    \includegraphics[clip,width=.8\columnwidth]{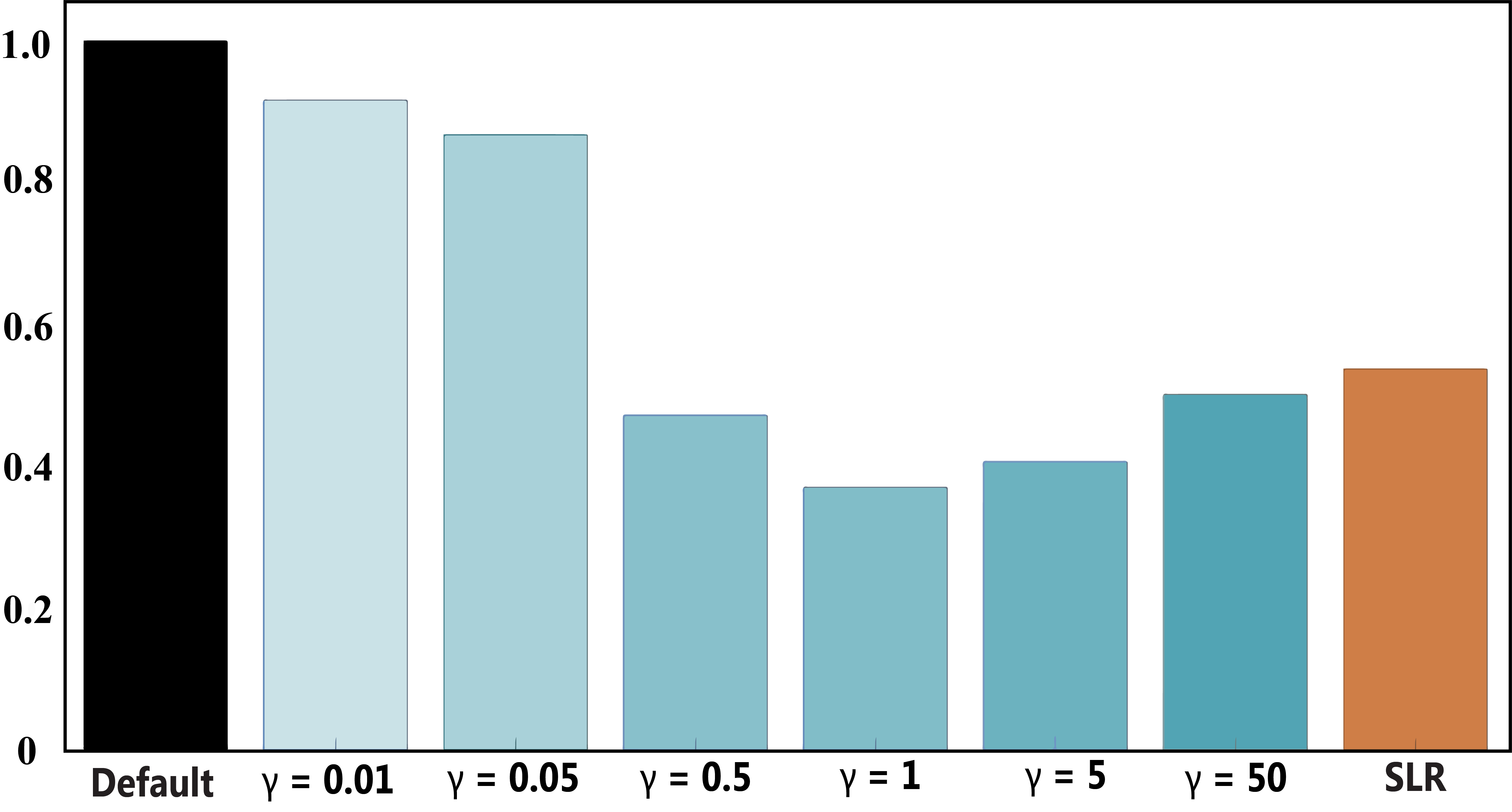}
    \caption{Overall (normalized) trajectory cost for variations in $\gamma$.}
\label{fig:liveness_ctrl_varaitions_in_gamma}
    \vspace{0em}
\end{figure}

\textit{Effect of $\gamma$}. 
To illustrate the effect of $\gamma$, we plot the overall trajectory cost \hl{(given by \eqref{eqn:mpc_cost_liveness}), normalized by the cost of the default controller} for different values of $\gamma$ in Fig. \ref{fig:liveness_ctrl_varaitions_in_gamma}. 
As evident from the figure, the choice of $\gamma$ can significantly affect the performance of the smooth blending filter.
Specifically, as $\gamma$ approaches 
$0$, the system behaves more and more like $\controller^*_{\text{live}}$, keeping stringent liveness characteristics.
This liveness, however, comes at the cost of a drop in performance (the controller cost is very high for small $\gamma$).
On the other hand, as we increase $\gamma$ (till around $\gamma = 1$), we see a proclivity towards using a controller \hl{that} is close to the nominal controller while encouraging liveness. 
This results in a significant decrease in the trajectory cost.
However, beyond a certain $\gamma$ ($\gamma = 1$ in this case), the cost is again seen to rise slightly. 
This is because, for such a high $\gamma$, the controller is allowed to behave more and more like the smooth LR filter.
The blending filter chooses the nominal controller as long as the system remains live; however, this choice leads the system closer to the BRT boundary, and the controller has to revert to a high-cost control in order to keep the system live, ultimately raising the overall controller cost. 
Hence, for high $\gamma$, the smooth-blending controller \hl{has} a similar cost as the smooth LR controller.
%
%
\subsection{Comparison of Different Liveness Filters}
\begin{figure}[t]
    \centering \includegraphics[width=0.7\columnwidth]{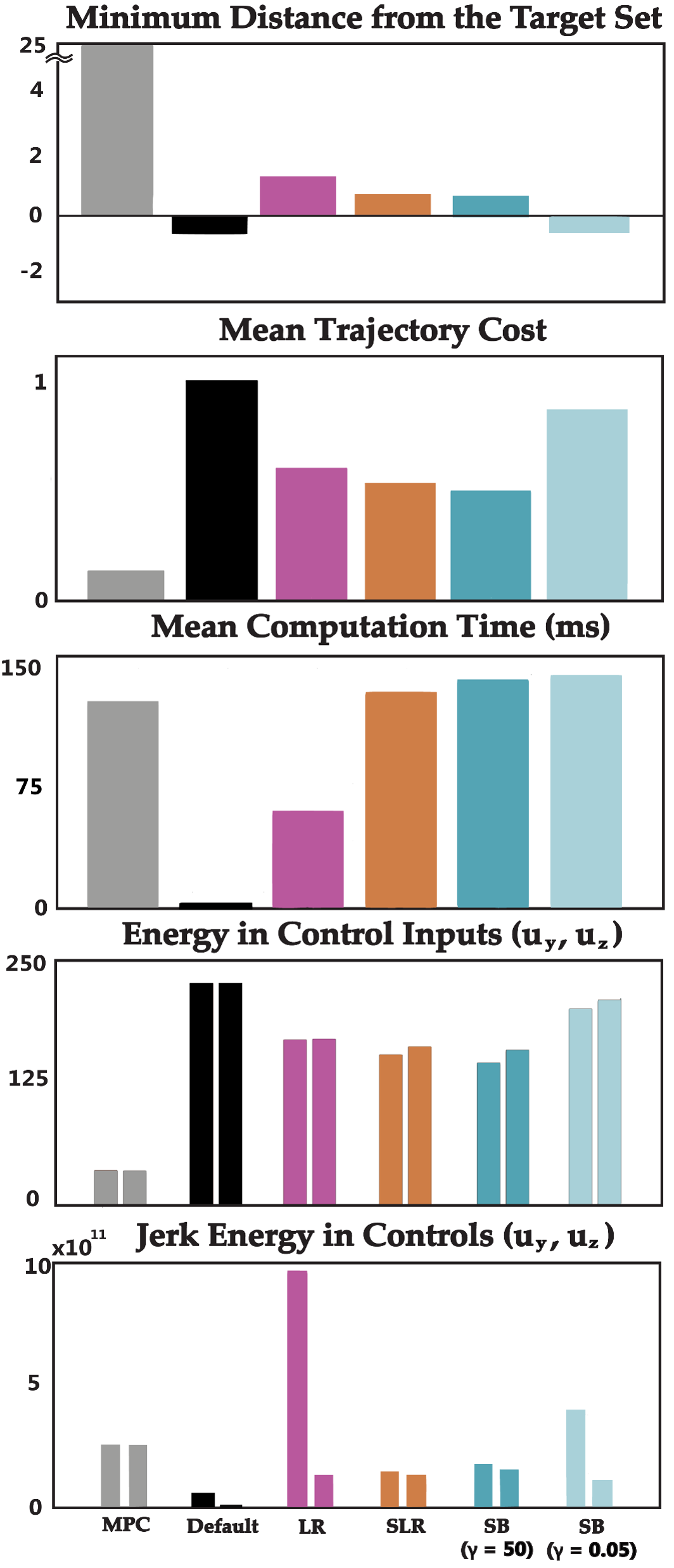}
    \caption{Comparison of different liveness filters across performance, computation time, and smoothness of the control profile. MPC is the nominal controller. LR and SLR stand for the least-restrictive and the smooth least-restrictive filters.
    SB stands for smooth blending.
}
    \label{fig:live_ctrl_comparisons}
    \vspace{0em}
\end{figure}

We now compare the performance of different liveness filters across five metrics: 
\begin{enumerate}
    \item \textit{minimum distance to the target set}, measuring the liveness of each filter;
    \item \textit{total cost} accumulated over the trajectory, measuring the performance of each filter;
    \item \textit{computation time} per step, measuring the filter latency;
    \item \textit{control energy}, measuring control authority used;
    \item \textit{jerk energy}, measuring the jerk in the control profile.
\end{enumerate}
The obtained results are summarized in \hl{Fig. \ref{fig:live_ctrl_comparisons}.}
%
%

%
%
%
%
First, we note that despite a poor nominal controller, all liveness filters steer the system very close to the target set (minimum distance to the target close to zero).
However, most filters have a small non-zero distance to the target set.
This is because in our simulation we use a small but non-zero timestep, which introduces a small delay in the application of the filtered policy, resulting in the system state being outside of the BRT momentarily.
We discuss this aspect further, as well as potential remedies in Sec. \ref{subsec:finite_timestep}. 
The exception is the smooth blending filter with small  $\gamma$, as it very closely behaves like the default controller sacrificing performance almost entirely to reach the target. 
%
In terms of performance, we note that $\controller^*_{\text{live}}$ results in a very high trajectory cost, mostly due a high control energy cost expected of its bang-bang nature.
In contrast, liveness filters balance liveness with performance, resulting in a lower trajectory cost.
In this case, \hl{we observe that smoother control profiles typically equate to lower trajectory costs}, resulting in a better performance by smooth blending filter followed by smooth LR filter followed by LR filter.
We further note that for $\gamma = 50$, the performance of the smooth blending and smooth LR filters is very close, as expected.
However, as $\gamma$ decreases, the smooth blending filter prioritizes liveness over performance, resulting in an increased trajectory cost.
We see a very similar pattern in the control energy plots.
%

Next, we compare the computation time of different controllers.
Here, the default controller emerges as the fastest option, as it only requires computing 
$\controller_{\text{live}}^{*}$, which typically is very fast, especially when the value function is represented as a neural network.
In fact, in this case, computing the nominal controller is more expensive than computing $\controller_{\text{live}}^{*}$.
For the same reason, the LR filter is faster than the MPC method, as it resorts to $\controller_{\text{live}}^{*}$ when the system state is at the BRT boundary.
Compared to the aforementioned controllers, the smooth LR and smooth blending filters leads to a higher computation time, as they require solving an optimization problem additionally to computing the nominal MPC controller. 
Moreover, the smooth LR filter has a lower computation cost among the two because it only solves an optimization problem when the system state is at the BRT boundary.
In general, the better the nominal controller is at maintaining liveness, the smaller the computation time for the smooth LR filter.

Finally, we measure the average jerk in the control trajectories generated by different filters. 
Despite its extremum seeking behavior, the default controller in this case has a low jerk, as the controller does not switch between the two extremes often.
In contrast, the LR filter leads to a sudden switch from the nominal controller to $\controller_{\text{live}}^{*}$ at the BRT boundary, leading to a chattering and high jerk.
The smooth LR and smooth blending filters smooth out these transitions between the two controllers, leading to lower jerks. 

\noindent \textbf{Overall recommendation.} Upon comparing different liveness filters, we recommend to use the smooth blending controller with a high value of $\gamma$, as it nicely trades off performance and liveness trough the tuning of $\gamma$, while maintaining a low jerk. 
At the same time, it allows to pick a large $\gamma$ to spare the user from tuning while keeping a considerable performance increase as seen in Fig. \ref{fig:liveness_ctrl_varaitions_in_gamma}.
A close second choice is the smooth least restrictive filter, which also shares the above advantages.
In addition, it has a lower computation latency, making it particularly suitable for real-world robotic systems. 
%
%

\section{Safety Filtering Using Reachability Analysis} \label{sec:safety}
We now turn our attention to safety filtering. One important difference to note is that the value function often converges for the safety problem, i.e. $\brtunsafe$ stops growing after some amount of time, beyond which the system has enough time to avoid the target set despite the worst case disturbance. 
Consequently, we can use the converged value function $V(\state)$, which is no longer a function of time $\tvar$, to synthesize safety controllers, and the resulting controllers are time-invariant and can be expressed as:
\begin{equation}
\label{eq:default_safety_ctrl}
\controller_{\text{safe}}^{*}(\state) =  \arg \max_{\ctrl \in \cset} \min_{\dstb \in \dset} \nabla \vfunc(\state) \cdot \dyn(\state, \ctrl, \dstb)
\end{equation}

Nevertheless, the proposed filters can easily be extended to incorporate time, similar to liveness filters. To demonstrate its effectiveness, we introduce the following example case.

\begin{figure} [t]
    \centering
    \includegraphics[width=0.9\columnwidth]{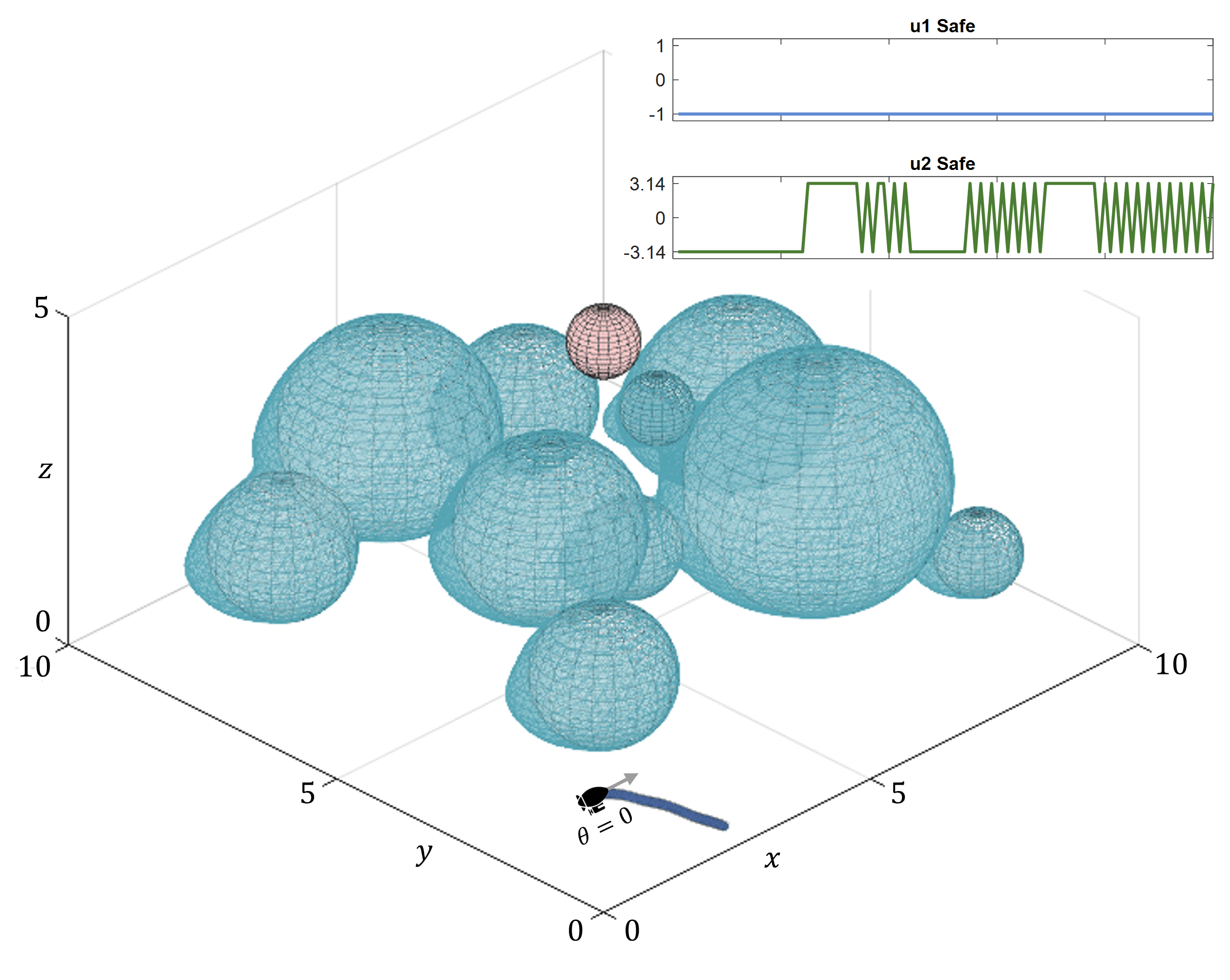}
    \caption{\hl{A $xyz$-slice (for $\theta=0$) of the BRT, $\brtunsafe$, for the safety example is shown in blue. 
    The gray spheres inside the BRT correspond to the obstacles (the target set in this case) and the pink sphere represents the set of goal states.}
    The system trajectory under the default safety controller is shown in dark blue and the corresponding controller profiles are shown in the inlets.
    }
    \label{fig:blimp_brt}
    \vspace{0em}
\end{figure}

\noindent \safetyexample
\hl{Consider a 4D system model of a blimp with state $\state=(x,y,z,\theta)$, with control over the rate of change of altitude and yaw angle, as described by the following dynamics:}
\begin{equation}\label{eq:blimp4D}
  \begin{aligned}
    &\dot{x}=v \cos (\theta) + {d}_{1} \\
    &\dot{y}=v \sin (\theta) + {d}_{2} \\
    &\dot{z}=u_{1}\\
    &\dot{\theta}=u_{2}
  \end{aligned}
\end{equation}
\hl{where $x,y$ and $z$ denote the $X,Y$ and $Z$ positions of the center of mass of the system, respectively, and $v$ is the speed of the system on the $XY$ plane.}
The control inputs are the vertical velocity in the $Z$ direction and the rate of change of heading.
In addition, there are velocity disturbances in the $X$ and $Y$ directions.
We consider a $10m$ x $10m$ x $5m$ position state space (Fig. \ref{fig:blimp_brt}) where the objective is to reach a target area of radius $R=0.5m$ located at  $(x,y,z)=(7,7,3)m$ (pink sphere) while avoiding collision with \hl{the obstacle set $\targetset$ that corresponds to a collection of spheres of various sizes spread across the position state space (gray spheres). In this case, the implicit obstacle function is defined as the signed distance to the closest gray sphere, $l(\state)=\min_i {d(\state, S_i)}$ where $d(\state, S_i)=\sqrt{(x_i-x)^2 +(y_i-y)^2 + (z_i-z)^2} - R_i$ is the distance from state $\state$ to the sphere $S_i$ of radius $R_i$ centered at $(x_i,y_i,z_i)$}. In (\ref{eq:blimp4D}) $v=2 m/s$ is the velocity in the $XY$ plane, the control and disturbance bounds are given as $u_1 \in [-1,1]m/s$, $u_2 \in [-\pi,\pi]\text{rad}/s$ and $||\begin{bmatrix}
    d_1 & d_2
\end{bmatrix}^\top||_2 \leq 0.5 m/s$.

For this example, \hl{to illustrate that the proposed method is agnostic to how the value function is obtained}, we leverage the Level Set Toolbox \cite{mitchell2007toolbox} and its reachability helper library HelperOC \cite{helperOC} to compute a numerical approximation of the BRT. 
Level-Set methods solve the HJB-VI numerically over a uniformly discretized state space grid. 
We use a 4D grid of $81 \times 81 \times 41 \times 21$ points for our computations. 
The solution of the HJB-VI provides us with the value function, $\vfunc(\state, \tdummy)$, which can be used to extract the $\brtunsafe$ similar to  \eqref{eqn:brt_live_value}.
In this case, the value function converges after 0.5 seconds; thus, we consider the time converged $\brtunsafe$. 
A 3D slice of $\brtunsafe$ for a fixed value of $\theta=0$ is presented in Fig.~\ref{fig:blimp_brt}.
The expansion of the BRT can be seen in the opposite direction, as states oriented this way will inevitably collide with obstacles if they start too close.
By definition of the BRT,  as long as the system starts outside the blue region, it is guaranteed to remain safe under the default safety controller $\controller_{\text{safe}}^{*}$ \eqref{eq:default_safety_ctrl}.
The system trajectory under $\controller_{\text{safe}}^{*}$ starting at a state outside the BRT and the corresponding control profiles are shown in Fig. \ref{fig:blimp_brt}.
The system remains safe under the default safety controller as expected; however, the default safety controller steers the system as far away from the obstacles as possible, disregarding any other performance criterion the system might have (such as reaching the pink goal region).
Moreover, the control profile is jittery, which is typically undesirable for real-world applications.
%
\subsection{Characterizing the set of safe controls}
Similar to liveness filtering, we will first characterize the set of \textit{all} safe control inputs for the system, and then use it to design various safety filters. 
To ensure safety, we would like the system to stay out of the $\brtunsafe$ at all times. 
Thus, intuitively, the set of safe control $\cset_{\text{safe}}(\state)$ at a state $\state$ are the control inputs that instantaneously keep the system state outside $\brtunsafe$, thereby ensuring recursive safety.
Mathematically,
%
\begin{equation} \label{eqn:safe_controls}
    \cset_{\text{safe}}(\state)=\left\{\begin{array}{l}
    \cset \quad \text { if } \vfunc(\state) > 0 \\
    \\
    \{\ctrl\in\cset: \nabla \vfunc(\state) \cdot \dyn(\state, \ctrl, \dstb^*) = 0 \} \\ \text { if } \vfunc(\state) = 0 \\
    \\
    \emptyset \quad \text { if } \vfunc(\state) < 0 \end{array}\right.
\end{equation}
%
The proof of \eqref{eqn:safe_controls} follows similarly to that for the liveness case, except that we do not need to account for the time-derivative of the value function in the controller expression, since the value function has converged.
The set of safe controls is $\cset$, when the system state is outside $\brtunsafe$.
When the system state is at the boundary of $\brtunsafe$, the set of safe controls are those that keep the state instantaneously at the boundary of $\brtunsafe$. 
Finally, safety cannot be ensured under any control input if the system state is inside $\brtunsafe$.
An explicit characterization of $\cset_{\text {safe}}$ suggests the following general structure of a safety filter:
\vspace{0.5em}
\begin{figure}[b]
    \centering    \includegraphics[width=1\columnwidth]{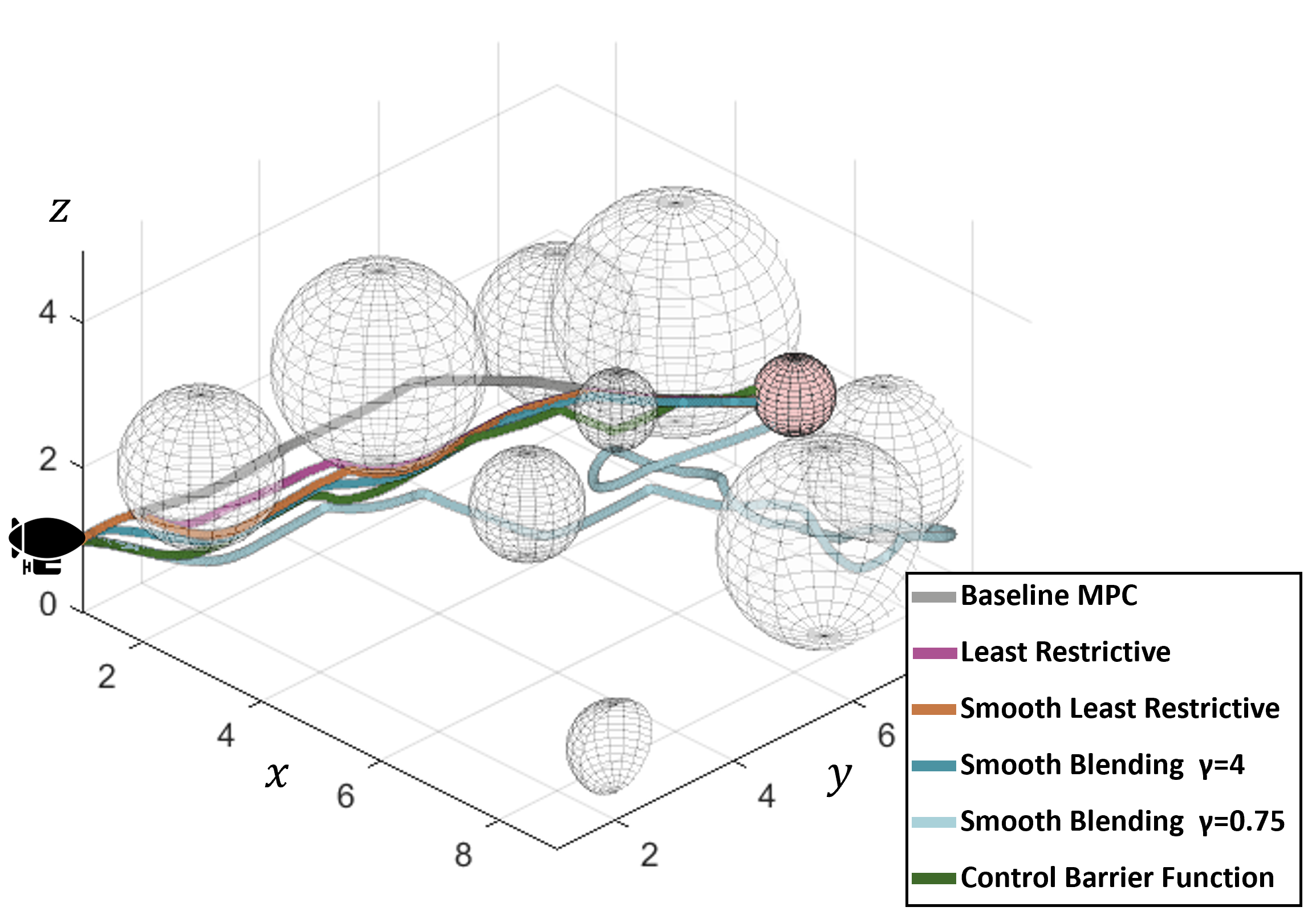}
    \caption{Trajectories for the nominal control, the proposed safety filters, \hl{and a CBF-based filter.}
    }
    \label{fig:safe_ctrl_traj}
    \vspace{0em}
\end{figure}
\noindent \textbf{General Safety Filter}. Given a nominal controller, $\ctrlnom$, a safety filter can be constructed as:
\begin{equation} \label{eqn:general_safety_filter}
    \begin{split} 
    \controller(\state, \tvar) = \arg\min_{\ctrl} & \quad h(\ctrl, \ctrlnom(\state, \tvar)) \\
    s.t. ~& \ctrl \in \widetilde{\cset}(\state, t), \text{~with~} \widetilde{\cset}(\state, t) \subseteq \cset_{\text{safe}}(\state)
\end{split}
\end{equation}
Similar to liveness filters, we refer to $h$ as the \textit{projection operator} and $\widetilde{\cset}$ as the \textit{projection set}.
Since $\widetilde{\cset} \subseteq \cset_{\text{safe}}$, the above filter makes sure that the system remains safe at all times.

\subsection{Least Restrictive Safety Filter} \label{sec:safety_filter_LR}
\begin{figure} [t]
    \centering
    \includegraphics[width=1\columnwidth]{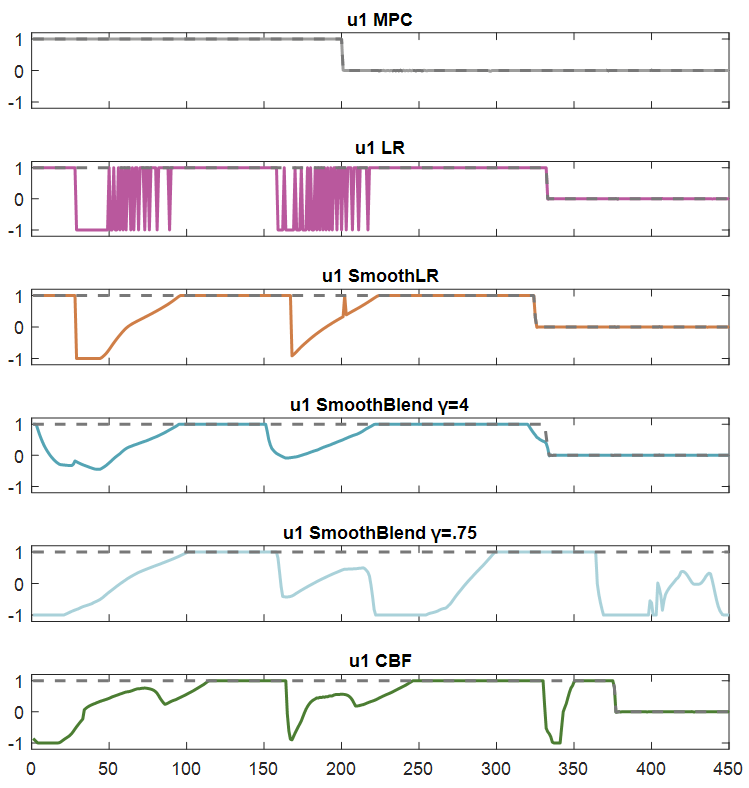}
    \caption{Control profiles for $u_1$ (z-velocity) for the nominal controller (gray dashed line), the proposed safety filters,  \hl{and a CBF-based filter.} 
    }
    \label{fig:safe_ctrl_profiles}
    \vspace{0em}
\end{figure}

%
The set of safe controls can be used to design a least restrictive safety controller for the system:
\begin{equation}\label{lst_restrict_safety_ctrl}
\controller(\state, \tvar) = \begin{cases}
  \ctrlnom(\state, \tvar) & \vfunc(\state)> 0 \\
   \controller_{\text{safe}}^*(\state) & \vfunc(\state) = 0
\end{cases}
\end{equation}
Similar to the liveness case, the least restrictive safety controller follows the nominal controller when the system is not risk of breaching safety, and it switches to $\controller_{\text{safe}}^{*}(\state)$ \eqref{eq:default_safety_ctrl} when the system is on the boundary of the BRT. 
The LR safety filter in \eqref{lst_restrict_safety_ctrl} can be obtained using our general framework in \eqref{eqn:general_safety_filter} by using the following $\widetilde{\cset}$:
\begin{equation}
\widetilde{\cset}(\state, \tvar) = \begin{cases}
  \cset&\quad \text { if } \vfunc(\state) > 0 \\
   \{\controller_{\text{safe}}^*(\state)\} & \quad \text { if } \vfunc(\state) = 0
\end{cases}
\end{equation}
\noindent \safetyexample
For this running example we use an MPC-based nominal controller, whose objective is to minimize distance between the system and the target area, subjected to the system dynamics and the control bounds. 
To highlight the impacts of safety filtering, the nominal controller is built such that it does not consider obstacle avoidance in its objective.
The system state trajectories and control profiles under the nominal controller are shown in grey in Fig. \ref{fig:safe_ctrl_traj} and Fig. \ref{fig:safe_ctrl_profiles} respectively.
As expected, the nominal controller fails to satisfy the safety constraint (i.e. avoiding the obstacles).
 
 We next demonstrate the least restrictive safety filter acting over the nominal controller. The corresponding system trajectory and control profiles (for $u_1$) are shown in Fig. \ref{fig:safe_ctrl_traj} (purple) and Fig. \ref{fig:safe_ctrl_profiles} respectively. 
This filtering technique allows the system to avoid collision with obstacles, but it comes with the shortcoming of an overly aggressive control profile at the boundary of the unsafe set near the obstacles, as the controller switching and the bang-bang nature of $\controller_{\text{safe}}^*$ results in a jittery control profile that jumps to the limits of the control authority.

%
\subsection{Smooth Least Restrictive Safety Filter}
%
Some of the shortcomings of the least restrictive filter can be addressed by the smooth least restrictive filter:
\begin{equation}\label{qp_safety_bd}
\controller(\state, \tvar) = \begin{cases}
      \ctrlnom(\state, \tvar)& \vfunc(\state)> 0 \\
       \controller^{+}(\state, \tvar) & \vfunc(\state) =  0
    \end{cases}
\end{equation}
where $\controller^{+}(\state)$ is obtained by solving the following optimization problem:
\begin{equation}\label{qp_safety_bd_opt}
\begin{split} 
    \min_{\ctrl \in \cset} & \quad ||\ctrl - \ctrlnom(\state, \tvar)||_2^2 \\
    s.t.     & \quad \nabla \vfunc(\state) \cdot \dyn(\state, \ctrl, \dstb^*) = 0
\end{split}
\end{equation}
The smooth least restrictive filter can readily be obtained using our general filtering framework by selecting $\widetilde{\cset} = \cset_{\text{safe}}$, i.e., using the maximal safe control set at all times.

Compared to the least restrictive filter, the smooth least restrictive filter allows the system to take a control action that is ``closest" to the nominal control, yet keeping it safe at all times, leading to a smoother control profile. 
Note that we have dropped the \hl{$D_{\tvar}$} term from the constraint in \eqref{qp_safety_bd_opt} because we assume the usage of the converged value function. 

\vspace{0.3em}
\noindent \safetyexample
%
%
The results of applying a smooth least restrictive filter to the running example is shown in orange in Fig. \ref{fig:safe_ctrl_traj} and  its control profiles in Fig. \ref{fig:safe_ctrl_profiles}. 
Wherever possible, the controller avoids the bang-bang nature of $\controller_{\text{safe}}^*$ by allowing the selection of controls that are close to the nominal controller, yet guaranteed to be safe. 
This smooths out an overly aggressive control profile generated by the least restrictive filter at the boundary of $\brtunsafe$.

\subsection{Smooth Blending of Performance and Safety}
Again quite similar to its liveness counterpart, the smooth blending safety filter utilizes the CBF-like constraint to blend the safety and performance objectives.
\begin{equation}
\begin{split} \label{qp_safety_smooth}
    \min_{\ctrl \in \cset} & \quad h(\ctrl, \ctrlnom(\state, \tvar)) = ||\ctrl - \ctrlnom(\state, \tvar)||_2^2 \\
    s.t.   & \quad \nabla \vfunc(\state) \cdot f(\state,u,d^*) \geq - \gamma \vfunc(\state)  
\end{split}
\end{equation}
Similar to the least restrictive and smooth least restrictive filters, the smooth blending filter allows the system state to move towards the boundary of $\brtunsafe$ (i.e., the value function can decrease), but unlike the aforementioned controllers, it limits the rate (determined by the user-defined coefficient $\gamma$ and the value of the current state $\state$) at which the value function is allowed to decrease.
This phenomenon (a) avoids a sudden switching to $\controller_{\text{safe}}^*(\state)$, leading to a smoother profile; and, (b) encourages the system to maintain a non-zero distance from the BRT boundary at all times.

The smooth blending filter can be obtained by using $\widetilde{\cset}(\state, t) = \{u \in \cset: \nabla \vfunc(\state) \cdot \dyn(\state, \ctrl, \dstb^*) \geq -\gamma V(\state)\}$ within the proposed general safety filter framework.
Similar to Lemma \ref{lemma:feasibility_smooth_blending}, it can be shown that $\widetilde{\cset}(\state, t) \subseteq \cset_{\text{safe}}(\state)$ for all $\gamma \geq 0$.

\vspace{0.3em}
\noindent \safetyexample
%
%
Resulting state trajectories of applying the smooth blending filter for two different values of $\gamma$ are shown in blue and light blue in Fig. \ref{fig:safe_ctrl_traj}, and their control profiles are shown in Fig. \ref{fig:safe_ctrl_profiles}. As evident from the control profiles, the smooth blending controllers do not produce any sudden switching since they are free to select controls closer to the nominal controls as long as the decrease in value is bounded by the prescribed rate. 
This rate is adjusted through the parameter $\gamma$, which determines how ``confident'' the system feels moving toward the unsafe states at a given value level. 
For this example we used values of $\gamma=4$ and $\gamma=0.75$. 
A larger value of $\gamma$ allows the system trajectory to move closer to the obstacles (increasingly behaving like a smooth LR filter), while a smaller value results in a more conservative behavior (behaving similar to the default safety controller).
This can also be seen from the system trajectory corresponding to $\gamma=0.75$, wherein the safety filter is pushing the system away from the obstacles to ensure that the value function doesn't decrease too much (similar to the default safety controller), resulting in a very curvy and inefficient trajectory to the goal. 
In contrast, the trajectory corresponding to $\gamma=4$ ventures fairly close to the BRT boundary. 
These trajectories highlight a critical characteristic of smooth blending filter -- the system performance can heavily depend on $\gamma$.
\hl{
Thus, a proper tuning of $\gamma$ is needed to obtain a satisfactory behavior.} 

\hl{Even though the proposed filter ensures safety for any value of $\gamma \geq 0$, we do not study mechanisms to compute an appropriate $\gamma$ in this work. 
An appropriate value of $\gamma$ will depend on how the user characterizes performance, safety conservativeness, and the compromise between them. The choice of gamma further depends on the system dynamics, failure set, control and disturbance bounds, and the nominal controller. 
For our running example, we simulated various values of $\gamma$ and picked two values to highlight the extreme behaviors that $\gamma$ can cause in the filtered trajectory.
}

\subsection{Comparison of Different Controllers}
\begin{figure}[t]
    \centering    \includegraphics[width=0.95\columnwidth]{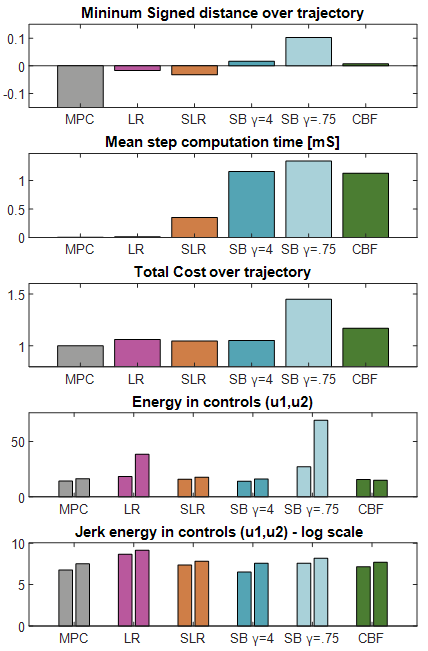}
    \caption{Comparison between the baseline MPC controller, proposed safety filters \hl{and a CBF-based filter.
    }}
    \label{fig:safe_comp}
    \vspace{0em}
\end{figure}

In this section, we present a comparison between the proposed filtering schemes across five metrics: \textit{total cost} accumulated over the trajectory, \textit{minimum distance to the obstacle set}, the controller \textit{computation time} per step, \textit{control energy}, and control \textit{jerk energy}.
The control profiles for $u_1$ are shown in Fig. \ref{fig:safe_ctrl_profiles} and the metrics for both $u_1$ and $u_2$ are presented in Fig. \ref{fig:safe_comp}.

The baseline MPC controller does not consider any obstacles, thus getting the most negative signed distance as it penetrates through them; for all other metrics, it will be considered as the base case we will use for comparison.
\begin{figure*}[ht!]
    \centering
    \begin{subfigure}{0.75\textwidth}
        \includegraphics[width=\textwidth]{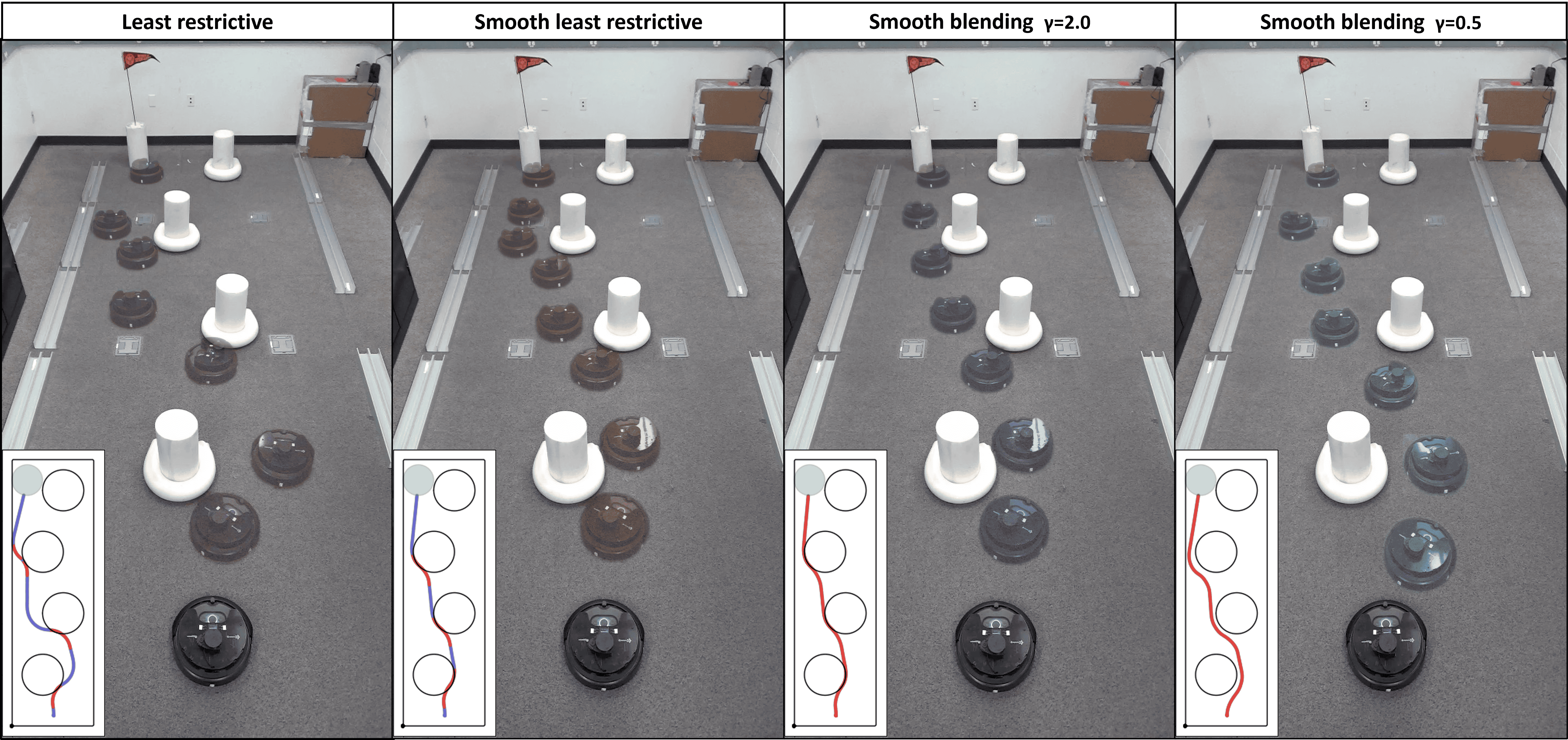}
    \end{subfigure}
    \begin{subfigure}{0.24\textwidth}
        \includegraphics[width=\textwidth]{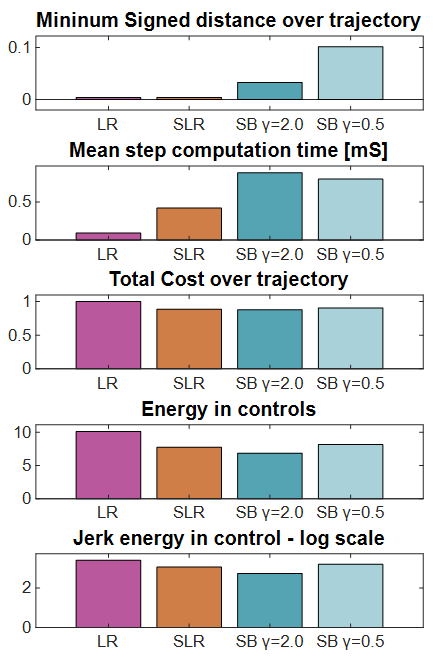}
    \end{subfigure}
    \caption{(Left) Trajectories for the Turtlebot safely navigating through the obstacle field to reach the goal area under different safety filters.
    A sampling-based MPC method is used as a nominal controller.
    Inlet plots consider the robot as a point entity and inflated obstacles, blue trajectories indicate the nominal controller is in operation, while red indicates that the safety filtered control is being applied to the robot.
    (Right) Comparison of different safety filters across performance, controller smoothness, computation time, and safety metrics.}
    \vspace{-1.0em}
    \label{fig:exp_main_result}
\end{figure*}

\hl{
Considering the smooth blending filter's CBF-inspired nature, we also implement a CBF-based safety filter applied to the same baseline MPC controller. Since the dynamical system defined by (\ref{eq:blimp4D}) is control-affine and the relative degree of $l(\state)$ to the dynamics is one, we can use the approach presented in \cite{HighOrderCBF} to use it as a High Order CBF. We choose the hyperparameters for this approach such that the CBF covers $\brtunsafe$, and the hyperparameter that corresponds to our parameter $\gamma$ is set to $4$ for a fair comparison.
}

The least restrictive filter maintains safety and only gets minimum penetration over obstacles; this slight penetration is due to the switching happening not strictly at $V(\state)=0$ but when the value function is negative due to finite grid and time step resolution.
The safety enforcement increases the total cost over the trajectory compared to the base MPC controller, because when safety is at risk, it steers the system away from unsafe states which conflicts with the baseline controller objective. 
The least restrictive filter is the fastest to implement among all safety filters as the default safety controller can be computed quickly using the BRT, without requiring to solve any optimization problem, unlike other safety filters. 
The downside lies in high energy and high jerk in the controls due to the bang-bang nature of the default safety controller.

The smooth least restrictive filter also suffers from some penetration over obstacles because it uses the same switching condition as the regular least restrictive filter, but presents improvements both in the energy and jerk of the controls as it no longer relies on purely bang-bang safety controller, with the drawback of increased computation time per step as \eqref{qp_safety_bd_opt} needs to be solved each time that safety is at risk.

The final three controllers correspond to smooth blending of performance and safety for two $\gamma$ values and the CBF-based filter.
Inclusion of the $-\gamma V(\state)$ term in the safety filter allows the system to reason about safety before reaching the boundary of the unsafe set, allowing the system to maintain a positive signed distance to obstacles over the whole trajectory.
This comes at the expense of an increased mean calculation time, especially for the case with smaller $\gamma$ as the more conservative constraint in \eqref{qp_safety_smooth} is slightly harder to solve.
\hl{The characteristics mentioned so far are shared with the CBF-based filter, where we observe very similar results to that of a smooth blending filter with $\gamma=4$. This is expected since we are essentially solving an equivalent optimization problem. We also observe some increased cost for the CBF-based filter, as it leads to a more conservative unsafe set compared to the BRT, which corresponds to the minimal unsafe set. The behavior of the CBF-based safety filter becomes even more conservative or overly optimistic if the hyperparameter during the CBF synthesis is not properly tuned. Thus, the safety and conservativeness of a CBF-based filter are directly affected by its synthesis, which can be challenging in practice for general nonlinear systems.}
It is also worth noting that a high $\gamma$ behaves quite similarly to the smooth LR filter, without its downside of accidentally penetrating the obstacle due to a last-minute switching.

\noindent \textbf{Overall recommendation.} Similar to the liveness case, we recommend to use the smooth blending safety filter with a high value of $\gamma$, as it nicely tradeoff performance and safety, while maintaining a low jerk. At the same time, using a very high $\gamma$ can bypass the need for tuning it.

\section{\label{cases}Hardware Experiments}
We next apply the proposed safety filters on a Turtlebot4, a Dubins-like robotic platform (see Fig. \ref{fig:exp_main_result}).
The robot needs to reach a goal position in an area situated with multiple obstacles. 
To complete this task, we use a shooting-based MPC nominal controller that does not consider obstacle avoidance in its objective.
We choose this nominal controller on purpose to highlight the effect of safety filtering.
The nominal controller will be filtered using the techniques presented in Section \ref{sec:safety} to guarantee safety while navigating towards the goal.
%
%

We model the dynamics of the system as a Dubins car with $V=0.3~m/s$ and $u \in [-0.75,0.75]~rad/s$ where the state $\state = [x,y,\varphi]^T$ represents the $x$ and $y$ positions and the heading of the robot:
\begin{equation} \label{eqn:dubins_dynamics}
\dot{x} =V \sin \varphi; \quad 
\dot{y} = V \cos \varphi; \quad
\dot{\varphi} =u
\end{equation}
The BRT for this experiment is calculated on a $[x,y,\varphi]$ grid of $[101; 281; 181]$ points, which gives a resolution of $2cm$ in position over the experiment space and $2^\circ$ in orientation.
The BRT was computed until convergence (in this case, the BRT converges after a horizon of $1s$). 
The converged value function is then used to synthesize various safety filters.

We assume perfect state measurement during our experiments.
The system state is obtained through the robot's internal pose estimation capability given by a combination of measurements from an IMU, optical floor tracking sensor, and wheel encoders. 
To avoid accidental breach of safety in least restrictive and smooth least restrictive filters due to limited grid resolution and discrete timestep (see Sec. \ref{sec:safety_filter_LR}), we use a non-zero threshold to activate the safety controller. 
Specifically, we use the condition $V(\state)=\epsilon$ (instead of $V(\state)=0$) to trigger switching to $\controller_{\text{safe}}^*$. 
A value of $\epsilon=0.05$ was used in the experimental results shown.

\begin{figure}[t]
    \centering
    \includegraphics[width=0.94\columnwidth]{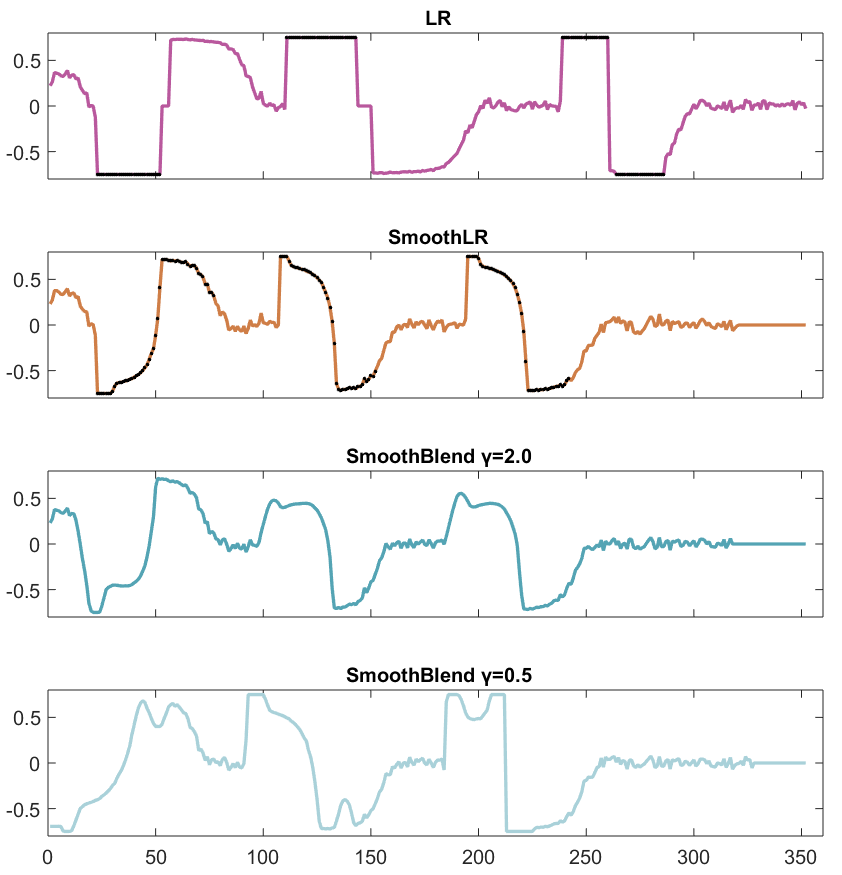}
    \vspace{0em}
    \caption{Control profiles for different safety filters obtained during our Turtlebot experiments. Black points in the Least Restrictive and Smooth Least Restrictive filters indicate that the filter is activated. They are omitted for the Smooth Blending filter as the filter is always active.}
    \label{fig:hw_safe_ctrl_profiles}
    \vspace{0em}
\end{figure}

The robot trajectories corresponding to different safety filters are shown in Fig. \ref{fig:exp_main_result}. 
The corresponding control trajectories are shown in \hl{Fig. \ref{fig:hw_safe_ctrl_profiles}}. 
We also compare different safety filters across various metrics. These results are shown in Fig. \ref{fig:exp_main_result} (right).
With this information, we highlight the core characteristics of each safety filter:

\textit{{Least Restrictive Filter}}: Even though the least restrictive filter keeps the system safe with fast queries of the default safety controller, the bang-bang nature of the safety controller forces the system to take the sharpest turn possible as it only reasons about maximally increasing safety. 
Additionally, it results in high energy control inputs.
Together, these results in a high accumulated cost over the robot trajectory.

\textit{{Smooth Least Restrictive Filter}}: Similarly to the previously presented simulation results, this method shows improvements compared to the least restrictive filter, both in the total energy and jerk of the control profile.
It also keeps lower accumulated cost as the safe control is trying to align with the underlying nominal controller. 
The drawback comes in the form of an increased computation time per step, as \eqref{qp_safety_bd_opt} must be solved each time the system safety is at risk.
It is also worth noting that using a non-zero threshold to \hl{activate} the QP-based safety controller in \eqref{qp_safety_bd} avoids any accidental collision with the obstacles, unlike our simulation results. 

\textit{{Smooth Blending of Performance and Safety}}: The final two trajectories show filtering with the smooth blending filter with $\gamma = 2$ and $\gamma = 0.5$.
The use of the filter on every time step of the trajectory (rather than just at the BRT boundary) allows the system to stay further away from the obstacles.
The drawback is the requirement to solve \eqref{qp_safety_smooth} on each step, which results in an increase in the mean step computation time.
The computation time, however, is still low enough to have any significant impact on robot operation.
We also observe that a smaller value of $\gamma$ leads to a more cautious robot behavior, reflected in the largest separation to obstacles among all safety filters (see Fig. \ref{fig:exp_main_result}). 
This comes at the cost of an increased control energy, control jerk, and an overall higher accumulated cost, as compared to $\gamma = 2$.

\section{ {Discussion of the Theory-Practice Gap}}
\label{sec:practical}
\subsection{Effect of a finite simulation timestep} \label{subsec:finite_timestep}
\begin{figure} [b]
    \centering
\includegraphics[width=.5\columnwidth]{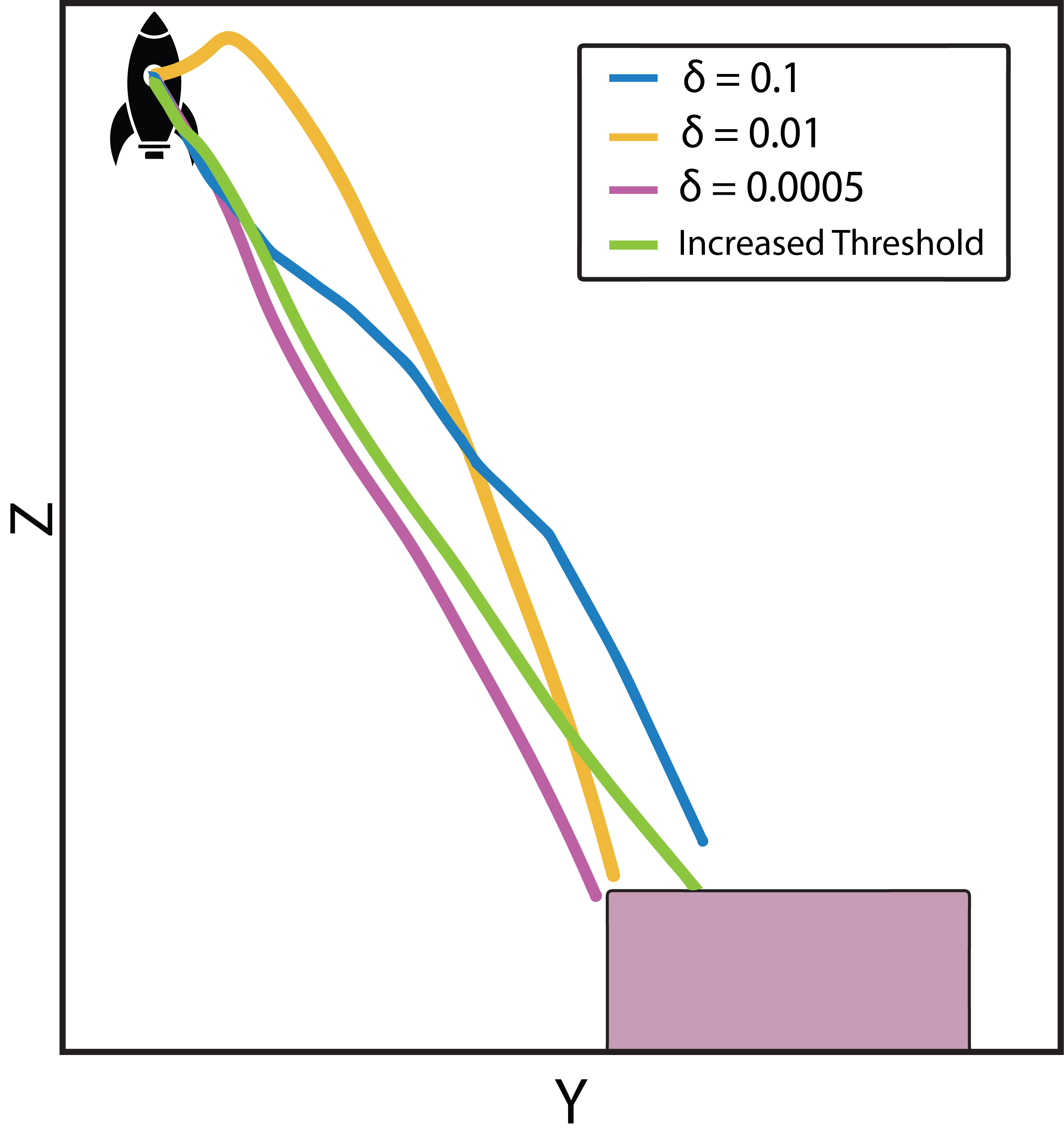}
    \caption{Effect of discretization timestep ($\delta$).  With all other parameters kept the same, the trajectories followed by the system under the same LR controller differ with different $\delta$ (blue $\delta = 0.1$, yellow $\delta = 0.10$, and pink $\delta = 0.0005$ trajectories). Using a modified version of the LR controller ($\delta = 0.0005, \epsilon = -0.1$), the system reaches the goal (green trajectory).
    }
    \label{fig:timestep_issue}
    \vspace{0em}
\end{figure}

%
%
The filters designed in this paper are based on the assumption that the control input can be applied to the system in a continuous time fashion. 
However, controller implementation on a real system often involves a zero-order hold, leading to a non-zero simulation timestep $\delta$.
If $\delta$ is too large, the proposed controllers may no longer be able to ensure liveness/safety for the underlying continuous time  system.
For example, \hl{a least restrictive filter} ensures liveness by switching to $\controller_{\text{live}}^*$ at the boundary of the BRT. 
However, \hl{under a finite $\delta$, $\controller_{\text{live}}^*$} may no longer be able to keep the system state within the BRT, as the liveness is only guaranteed under a \textit{continuous time}, state feedback reachability controller. 
This discrepancy between the continuous-discrete structure is demonstrated in Fig. \ref{fig:timestep_issue}, where we show the trajectory variations generated by the least restrictive filter for different values of $\delta$. 
%

Such effects are more pronounced in the least restrictive filters since there, the system evolves along the BRT boundary. If our simulation timestep is too high, it is highly possible that our system temporarily escape the BRT and enter the unsafe region. 
One possible solution could be a faster feedback with gradually decreasing $\delta$ as the system approaches the BRT or adaptive $\delta$ depending on the value function. 
One can also use a discrete-time variant of the BRT computation itself. 
Alternatively, instead of the switching at the BRT boundary ($\vfunc(\state) = 0$) as seen in Eqn. \eqref{lst_restrict_safety_ctrl}, we can choose to switch with some $\epsilon$ margin from the boundary. This slightly modifies the least restrictive controller as follows,
\begin{equation}\label{lst_restrict_liveness_ctrl_mod}
\controller(\state, \tvar) = \begin{cases}
  \ctrlnom(\state, \tvar)&\quad \text { if } \vfunc(\state, \tvar)< \epsilon \\
   \controller_{\text{live}}^*(\state, \tvar) & \quad \text { if } \vfunc(\state) \geq \epsilon
\end{cases}
\end{equation}
An example corresponding to $\epsilon = -0.1, \delta = 0.0005$ is shown with the green line in Fig. \ref{fig:timestep_issue}.
As evident from the figure, this strategy is able to maintain liveness at all times and eventually reaches the target set (unlike the pink trajectory using the same $\delta = 0.0005$, \hl{but $\epsilon = 0$)}.
This strategy was also employed during our hardware experiments to avoid accidental collisions with the obstacles.
Nevertheless, \hl{a thorough analysis of the effect of a finite simulation timestep} on system liveness/safety would be a promising future research direction.

\subsection{Non-differentiable value functions}
The proposed controllers in this paper rely on differentiability of the value function, which may not always be the case \cite{evans1984differential, mitchell2005time}. 
To discuss the impact of non-differentiability of the value function on the proposed controllers, consider the Dubins car dynamics in \eqref{eqn:dubins_dynamics}. 
Moreover, we consider the smooth least restrictive safety filter for our illustration.
This requires solving an optimization problem where we find the closest control to the nominal control, such that the condition $\langle \nabla \vfunc(\state), f(\state,u,d^*) \rangle = 0$ with $\ctrl \in \cset$ is met.

\begin{figure} [t]
    \centering
    \includegraphics[width=1.0\columnwidth]{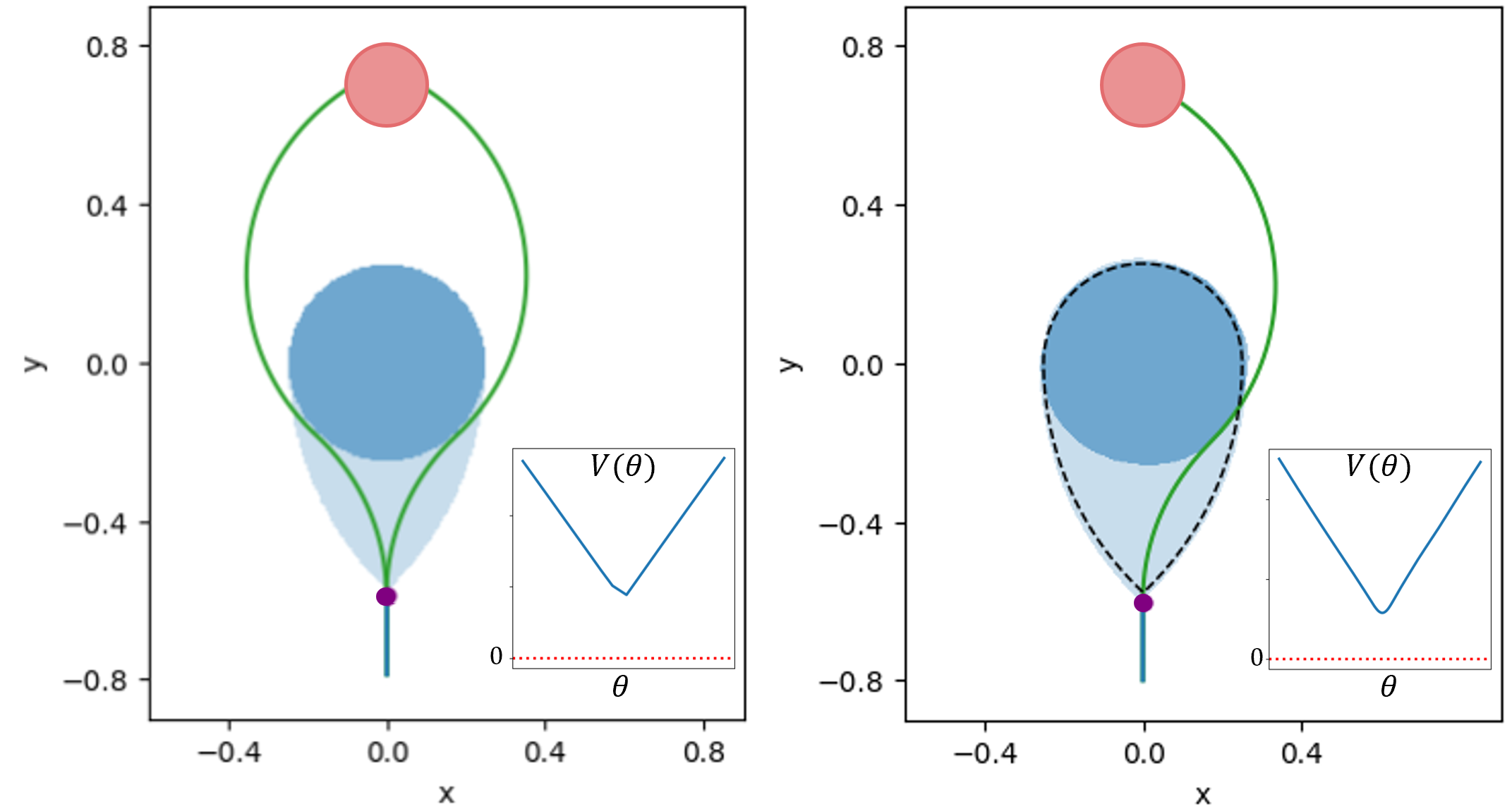}
    \caption{
    (Left) BRT slice for $\theta=\pi/2$. The value function computed using a grid-based approach exhibits non-differentiability at the purple point. Inlet: the value function plot at the purple point as a function of theta. The value function exhibits a kink at $\theta=\pi/2$.
    The green trajectories are computed using one-sided gradients. 
    \hl{(Right) A smooth approximation of the value function computed using a learning-based method (the black dashed line is the BRT corresponding to the grid-based approach). The value function now changes smoothly as a function of $\theta$ at the purple point.}}
    \label{fig:grad issue}
    \vspace{0em}
\end{figure}

\hl{First we consider the scenario where the value function is being numerically calculated using a grid based approach~\cite{helperOC},\cite{mitchell2007toolbox}.}
The goal region and the obstacle are shown in light pink and gray respectively in Fig. \ref{fig:grad issue}(left).
The nominal planner, unaware of the environment obstacle, plans a straight path to the goal region (blue trajectory).

Correspondingly, the safety controller engages when the system state reaches the BRT boundary (purple point). 
The value function has a kink at this state, wherein the system has two equally viable option (hard left-turn  or hard right-turn) to barely avoid the obstacle.
However, at this state, the value function gradient in the theta direction does not exist, as is evident from the inlet in Fig. \ref{fig:grad issue}(left) showing the variations on value around this point as a function of $\theta$ .
Thus, one can no longer use the safety filters that rely on the value function gradient (such as a smooth least restrictive filter or a smooth blending filter). 
%
One approach to avoid this issue could be to use one-sided gradient of the value function to construct the safety filter. 
The result of using a left-sided or a right-sided approximation of the gradient is shown in green trajectories in Fig.\ref{fig:grad issue}(left). Both approximations are able to keep the system safe while steering it to its goal location.


\hl{A second approach is to obtain a conservative approximation of the value function that is differentiable and then construct safety/liveness filters using this approximation.
For illustration, here we use a learning-based method, DeepReach, to approximate the value function using neural network representations that are differentiable by construction \cite{deepreach, linScenario}.
In Fig.\ref{fig:grad issue}(right), we show a comparison of the BRT obtained through DeepReach and the previously discussed grid-based method. 
The inlet in Fig.~\ref{fig:grad issue}(right) shows how the learned value function is differentiable at the tip of the BRT, unlike the grid-based value function. 
Moreover, the BRT obtained through the learning method encompasses the numerically approximated value function, which keeps all the safety guarantees intact at the expense of a slightly more conservative BRT, for completeness the smooth least restrictive filtered trajectory using this differentiable representation is shown in green.}

\section{Conclusion And Future Work} \label{sec:conclusion}
%
In this paper, we explicitly characterize the set of all live and safe controls for a dynamical system using Hamilton-Jacobi reachability analysis.
Leveraging these maximal sets, we introduce a general framework for designing safety and liveness filters for the system. 
We propose three distinct safety/liveness filters, balancing performance, control smoothness, and latency, allowing to select based on system needs.

While these filters integrate safety/liveness and performance, they modify the nominal controller only at the current timestep, unaware of the long-term effects of its action. 
This might result in myopic controllers.
Second, the discrete time nature of actual hardware implementations of these controllers might pose a challenge to the guarantees proposed. 
\hl{Third, we currently assume that the target set is known and does not change online. However, that may not be the case for many robotics applications and an online adaptation of the reachability analysis and safety filters might be required.}
We aim to tackle these challenges in future works.
\hl{Finally, we currently rely on the differentiability of the value function to construct safety and liveness filters. 
A theoretical analysis of filter design using subgradients or one-sided gradients of the value function is another important future research direction.}


\bibliographystyle{plain}
\bibliography{CDC2023, reachability} 

\appendix
\subsection{Proof of \textit{Lemma} \ref{lemma:u_live_set}}
\noindent {\textit{Case 1: $V(\state, t)<0$.}}
    Since the value function is continuous in state and time, and the system trajectory itself is continuous in time, the value function is continuous along the system trajectory. 
    Thus, we can always pick a small enough $\delta$ such that $V\left(\xi_{\state, t}^{u, d^*}(t+\delta), t+\delta\right) <0$. 
    That is, we can keep the system momentarily inside the BRT for any $u$.
    Thus, $\cset_{\text {live }}(\state, t) \equiv \cset$.

    \vspace{1em}
\noindent \textit{Case-2: $V\left(\state, t\right)=0$.}
    Since by assumption the value function is differentiable for all $\state$ and $\tvar$, for sufficiently small $\delta > 0$, we can rewrite the value function using Taylor expansion:
    \begin{align*}
        & V\left(\traj_{\state, t}^{u, d^*}(t+\delta), t+\delta \right) \\
        & \approx V\left(\state, t\right)+\frac{\partial V}{\partial t} \delta+\frac{\partial V}{\partial \state} \cdot f\left(\state, u, d^*\right) \delta \\
        & = \delta \left[ \frac{\partial V}{\partial t} + \frac{\partial V}{\partial \state} \cdot f\left(\state, u, d^*\right) \right]
    \end{align*}
    where the equality follows because $V\left(\state, t\right)=0$.
    Thus, to ensure that $V\left(\xi_{\state, t}^{u, d^*}(t+\delta), t+\delta\right) \leq 0$, we must have 
    \begin{equation*}
        \left[ \frac{\partial V}{\partial t} + \frac{\partial V}{\partial \state} \cdot f\left(\state, u, d^*\right) \right] \leq 0
    \end{equation*}
    Since $\state \notin \targetset$, $\targetfunc(\state) > 0$. It follows immediately that $\targetfunc(\state) - \vfunc(\state,\tvar) = \targetfunc(\state) - 0 > 0$.
    Thus, according to the HJI-VI in \eqref{eq:HJIVI_BRS}, we must have:
    $$
    \begin{aligned}
    & \frac{\partial V}{\partial t}+\min _u \frac{\partial V}{\partial \state} \cdot f\left(\state, u, d^*\right)=0 \\
    & \equiv \frac{\partial V}{\partial t}+\frac{\partial V}{\partial \state} \cdot f\left(\state, u, d^*\right) \geqslant 0 \quad \forall u \in \cset
    \end{aligned}
    $$
    Thus, the only feasibility for ensuring liveness is 
    \begin{equation*}
        \left[ \frac{\partial V}{\partial t} + \frac{\partial V}{\partial \state} \cdot f\left(\state, u, d^*\right) \right] = 0
    \end{equation*}
    which corresponds to $V\left(\xi_{\state, t}^{u, d^*}(t+\delta), t+\delta\right) = 0$
    
    \vspace{1em}
\noindent  \textit{Case-3: $V\left(\state, t\right) > 0$.} 
    In this case, the set of live controls is trivially an empty set, as per the definition of a BRT. 

\subsection{Proof of \textit{Corollary} \ref{cor:brt_live_u_live}} 
    
Suppose the value function $\vfunc(\state, \tvar)$ is differentiable at all $\state$ and $\tvar$. Take $\state \in \brtlive$. We split into the following two cases. 

\noindent \textit{Case} 1: $\vfunc(\state, \tvar) < 0$. Since $\controller_{\text{live}}^{*}(\state,\tvar) \in \cset$, we also have $\controller_{\text{live}}^{*}(\state,\tvar) \in \cset_{\text{live}}(\state, \tvar)$ as  $\cset \equiv \cset_{\text{live}}(\state, \tvar)$ in this case.

\vspace{1em}
\noindent \textit{Case} 2: $\vfunc(\state, \tvar) = 0$.
According to the HJI-VI in \eqref{eq:HJIVI_BRS}, we must have:
$$
    \begin{aligned}
    & \frac{\partial V}{\partial t}+\min _u \frac{\partial V}{\partial \state} \cdot f\left(\state, u, d^*\right)=0
    \end{aligned}
$$
By the definition of the default controller, it achieves the minimum in the above equation, i.e.,
$$
    \begin{aligned}
    & \frac{\partial V}{\partial t}+ \frac{\partial V}{\partial \state} \cdot f\left(\state, \controller_{\text{live}}^{*}(\state,\tvar), d^*\right)=0
    \end{aligned}
$$
Thus, $\controller_{\text{live}}^{*}(\state,\tvar) \in \cset_{\text{live}}(\state, \tvar)$ by the definition of the set of live controls.

\subsection{Proof of \textit{Lemma} \ref{lemma:u_live_guarantee}}

We will prove this result by contradiction.
    Suppose the system never reaches the target set in the time interval $\left[t, T\right]$. Then we must have $V(\state(T), T)>0$ (if not, then $V(\state(T), T) = l(\state(T)) < 0$ and we are done).
    
    Since $V\left(\state(t), t\right)<0$ and the value function is continuous, we must have that $V(\state(\tilde{t}), \tilde{t})=0$ for some $\tilde{t} \in\left(t, T\right)$.
    Moreover, for state $\state(\tilde{t})$, the set of live controls is given by (Eq. \eqref{eqn:live_controls}
    $$
    \frac{\partial V}{\partial t}+\frac{\partial V}{\partial \state} \cdot f\left(\state, u, d^*\right)=0
    $$
    It follows immediately that $\frac{dV}{dt} = \frac{\partial V}{\partial t}+\frac{\partial V}{\partial \state} \cdot f\left(\state, u, d^*\right)=0$. Thus, we must have:
    $
    V(\state(\tau), \tau)=0 \quad \forall \tau \geqslant \tilde{t}
    $
    This contradicts our initial hypothesis that $V(\state(T), T)>0$. Hence, the system must reach the target set at some point in the time interval $\left[t, T\right]$.

\subsection{Proof of \textit{Lemma} \ref{lemma:feasibility_smooth_blending}}
\noindent \textit{Case 1:} $V(\state, t)=0$. 
    In this case, 
    \begin{equation*}
        \widetilde{\cset}(\state, t)=\{u \in \cset: D_{t} \vfunc(\state, t) + \nabla \vfunc(\state, t) \cdot \dyn(\state, \ctrl, \dstb^*) \leq 0\}
    \end{equation*}
    However, as per the HJI-VI in \eqref{eq:HJIVI_BRS}, we have 
    \begin{equation*}
        D_{t} \vfunc(\state, \tvar) + \nabla \vfunc(\state, \tvar) \cdot \dyn(\state, \ctrl, \dstb^*) \geq 0 ~\forall u \in \cset
    \end{equation*}
    Thus, $\widetilde{\cset}(\state, t)$ can be equivalently written as 
    \begin{align*}
        \widetilde{\cset}(\state, t) = & \{u \in \cset: D_{t} \vfunc(\state, t) + \nabla \vfunc(\state, t) \cdot \dyn(\state, \ctrl, \dstb^*) = 0\} \\
        = & ~\cset_{\text{live}}(\state, t)
    \end{align*}
    Moreover, since $\controller_{\text{live}}^{*}(\state,\tvar) \in \cset_{\text{live}}(\state, \tvar)$ by Corollary \ref{cor:brt_live_u_live},  we have that $\controller_{\text{live}}^{*}(\state,\tvar) \in \widetilde{\cset}(\state, t)$.
    Thus, $\widetilde{\cset}(\state, t)$ is non-empty.

    \vspace{1em}
    \noindent \textit{Case-2:} $V(\state, t)<0$.
    In this case, $\cset_{\text {live}}(\state, t) = \cset$ so $\widetilde{\cset}(\state, t) \subseteq \cset_{\text{live}}(\state, t)$ trivially. 
    
    To prove that $\widetilde{\cset}(\state, t)$ is non-empty, let $\alpha = -\gamma V(\state, t)$. 
    Thus, $\tilde{\cset}(\state, t) = \{u: D_{t} \vfunc(\state, t) + \nabla \vfunc(\state, t) \cdot \dyn(\state, \ctrl, \dstb^*) \leq \alpha\}$.
    Note that since $V(\state, t)<0, \alpha>0$.
    Moreover, since for the default liveness controller, we have that $D_{t} \vfunc(\state, t) + \nabla \vfunc(\state, t) \cdot \dyn(\state, \controller_{\text{live}}^{*}(\state,\tvar), \dstb^*) = 0 < \alpha$, $\controller_{\text{live}}^{*}(\state,\tvar) \in  \tilde{\cset}(\state, t)$.
    Thus, $\widetilde{\cset}(\state, t)$ is non-empty.
\newpage
\begin{IEEEbiography}[{\includegraphics[width=1in,height=1.25in,clip,keepaspectratio]{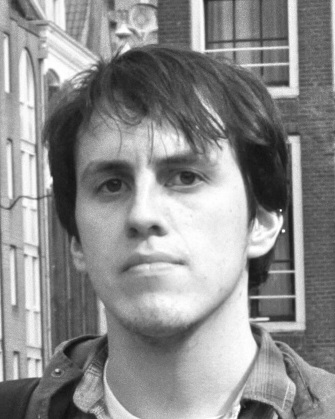}}]{Javier Borquez}
Currently pursuing a Ph.D. in Electrical and Computer Engineering at the University of Southern California, Los Angeles, as a member of the Safe and Intelligent Autonomy Lab. 
He received his M.S. and B.S. in Electrical Engineering from the University of Santiago de Chile in 2017 and 2015, respectively. His research interests lie in using optimal control theory, numerical methods, and learning-enabled approaches to develop safety-guaranteeing frameworks for the real-world deployment of robots.
\end{IEEEbiography}
\vskip -2\baselineskip plus -1fil

\begin{IEEEbiography}[{\includegraphics[width=1in,height=1.25in,clip,keepaspectratio]{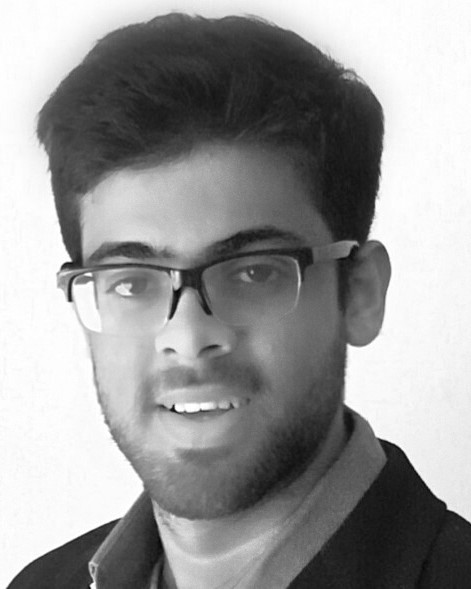}}]{Kaustav Chakraborty}
 Ph.D. student in the ECE department at the University of Southern California, Los Angeles, where he is advised by Prof. Somil Bansal and is a member of the Safe and Intelligent Autonomy Lab. Before USC, he received his M.S. in Robotics from the University of Michigan, Ann Arbor, and his B.Tech in Mechanical Engineering from Vellore Institute of Technology, India. His research centers on designing safety frameworks for autonomous systems using sensory feedback.
\end{IEEEbiography}
\vskip -2\baselineskip plus -1fil

\begin{IEEEbiography}[{\includegraphics[width=1in,height=1.25in,clip,keepaspectratio]{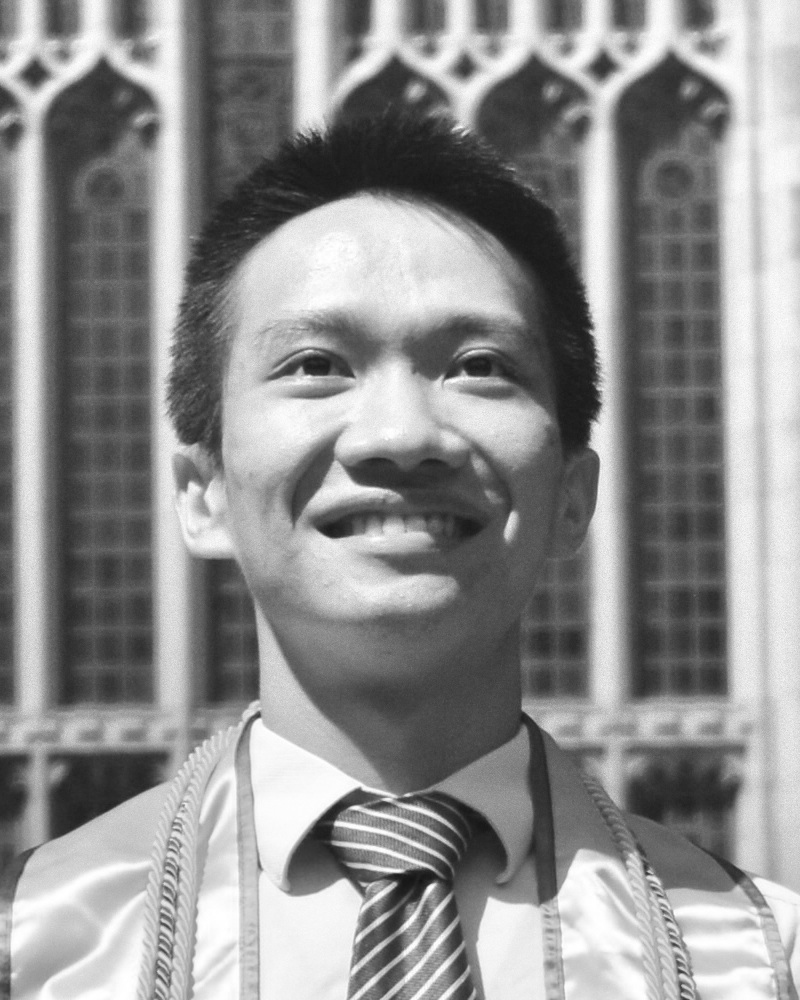}}]{Hao Wang}
Currently pursuing the Ph.D. degree in Electrical and Computer Engineering at the University of Southern California, Los Angeles. He received the B.S. degrees in computer science and mechanical engineering from the University of Michigan, Ann Arbor, in 2022. 
His research interests include optimal control theory and safety in autonomous systems. He is interested in utilizing tools from control theory, machine learning, and optimization to synthesize safe and performant controllers.
\end{IEEEbiography}
\vskip -2\baselineskip plus -1fil

\begin{IEEEbiography}[{\includegraphics[width=1in,height=1.25in,clip,keepaspectratio]{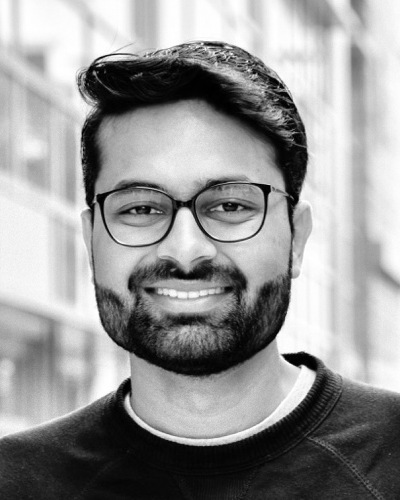}}]{Somil Bansal}
Assistant professor in the ECE department at the University of Southern California, where he leads the Safe and Intelligent Autonomy lab. He received a Ph.D. in Electrical Engineering and Computer Sciences from the University of California at Berkeley in 2020. Before that, he obtained a B.Tech. in Electrical Engineering from the Indian Institute of Technology, Kanpur, and an M.S. in Electrical Engineering and Computer Sciences from UC Berkeley in 2012 and 2014, respectively. His research focuses on understanding how machine learning methods can be combined with classical, model-based planning and control methods to enable intelligent and safe decision-making.
\end{IEEEbiography}

\end{document}